\documentclass[10pt,twocolumn,letterpaper]{article}

\usepackage[pagenumbers]{cvpr} % To force page numbers, e.g. for an arXiv version

\definecolor{cvprblue}{rgb}{0.21,0.49,0.74}
\usepackage[pagebackref,breaklinks,colorlinks,allcolors=cvprblue]{hyperref}
\usepackage{graphicx}
\usepackage{amsmath}
\usepackage{amssymb}
\usepackage{booktabs}
\usepackage{multirow}
\usepackage{tabu}
\usepackage{colortbl}
\usepackage{arydshln} 
\usepackage{stfloats}
\usepackage{makecell}
\usepackage{listings}
\usepackage{subcaption}
\usepackage{pifont}
\usepackage{adjustbox}
\usepackage{array}
\def\paperID{1393} % *** Enter the Paper ID here
\def\confName{CVPR}
\def\confYear{2025}
\usepackage[hang,flushmargin,symbol]{footmisc}

\title{LamRA: Large Multimodal Model as Your Advanced Retrieval Assistant}

\author{Yikun Liu$^{1,2*}$, Pingan Chen$^3$, Jiayin Cai$^3$, Xiaolong Jiang$^3$, Yao Hu$^3$,\\[2pt] Jiangchao Yao$^{2}$, Yanfeng Wang$^{1\dagger}$, Weidi Xie$^{1\dagger}$ \\[3pt]
$^{1}$School of Artificial Intelligence, Shanghai Jiao Tong University, China \hspace{0.5cm} \\
$^{2}$CMIC, Shanghai Jiao Tong University, China \hspace{0.5cm}
$^{3}$Xiaohongshu Inc., China \hspace{0.5cm}
}

\begin{document}

\twocolumn[{%
\renewcommand\twocolumn[1][]{#1}%
\maketitle
\begin{center}
   \centering
   \includegraphics[width=\textwidth]{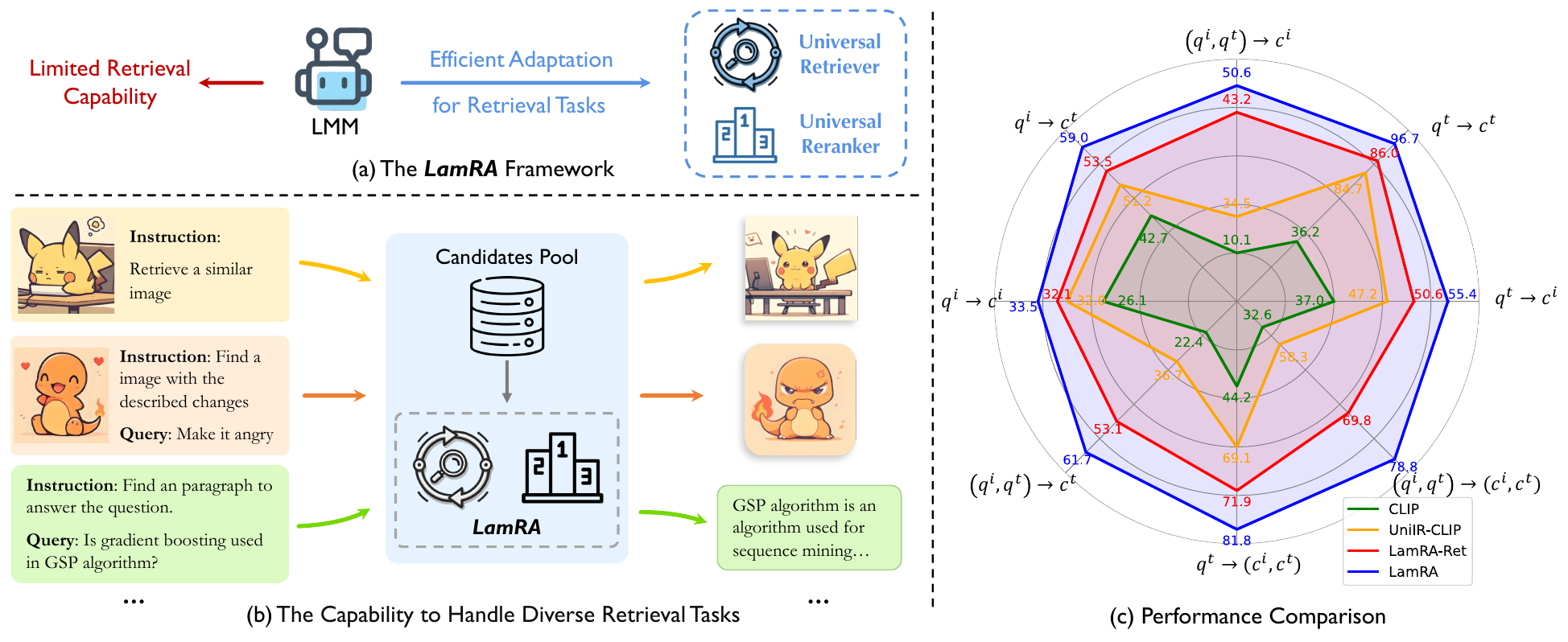} % for arXiv
   \captionof{figure}{The \textbf{LamRA} framework empowers Large Multimodal Models with advanced retrieval and reranking capabilities. (a) LamRA enhances LMMs with universal retrieval and reranking capabilities by inserting lightweight LoRA modules into the LMMs. (b) Examples of varied retrieval tasks demonstrate LamRA's capability to handle diverse retrieval tasks. (c) Performance comparison on the M-BEIR test set shows LamRA's superior performance across a wide range of retrieval tasks. For instance, $q^t \to c^i$ represents text-to-image retrieval.
   }
  \label{fig:teaser}
 \end{center}
 }]

 \renewcommand{\thefootnote}{\fnsymbol{footnote}} %将脚注符号设置为fnsymbol类型，即特殊符号表示
\footnotetext[1]{Work was done during internship in Xiaohongshu.}  
\footnotetext[2]{Corresponding author.}

% \blfootnote{
% *: work was done during internship in Xiaohongshu.\\
% $\dagger$: Corresponding author.
% }
% \begin{document}
% \maketitle

\begin{abstract}

With the rapid advancement of multimodal information retrieval, increasingly complex retrieval tasks have emerged. Existing methods predominately rely on task-specific fine-tuning of vision-language models, often those trained with image-text contrastive learning. 
In this paper, we explore the possibility of re-purposing generative Large Multimodal Models (LMMs) for retrieval. This approach enables unifying all retrieval tasks under the same formulation and, more importantly, allows for extrapolation towards unseen retrieval tasks without additional training. Our contributions can be summarised in the following aspects: 
(i) We introduce \textbf{LamRA}, a versatile framework designed to empower LMMs with sophisticated retrieval and reranking capabilities. 
(ii) For retrieval, we adopt a two-stage training strategy comprising language-only pre-training and multimodal instruction tuning to progressively enhance LMM's retrieval performance. 
(iii) For reranking, we employ joint training for both pointwise and listwise reranking, offering two distinct ways to further boost the retrieval performance. (iv) Extensive experimental results underscore the efficacy of our method in handling more than ten retrieval tasks, demonstrating robust performance in both supervised and zero-shot settings, including scenarios involving previously unseen retrieval tasks. Project page: \href{https://code-kunkun.github.io/LamRA/}{https://code-kunkun.github.io/LamRA/}. 
\vspace{-1.7em}
\end{abstract}

% \weidi{not impressive enough, give numbers, how many different retrieval tasks, datasets, blabla}
\section{Introduction}

% retrieval的现状和问题
\noindent Recently, the field of multimodal information retrieval has been significantly advanced by the remarkable success of Vision-Language Models (VLMs). For instance, CLIP~\cite{radford2021learning} and ALIGN~\cite{jia2021scaling}, which are trained through contrastive learning~\cite{oord2018representation} on large-scale image-text pairs, have demonstrated surprisingly impressive generalizability on cross-modal retrieval. However, as the landscape of information retrieval evolves, 
more challenging tasks have emerged, including composed image retrieval~\cite{liu2021image, wu2021fashion}, long-text image retrieval~\cite{zhang2024long}, and image/question to multimodal document retrieval~\cite{mensink2023encyclopedic, chen2023can}. Existing methods~\cite{liu2023zero, yan2024echosight, xusentence} employ task-specific fine-tuning to adapt VLMs to these specialized tasks, which remains cumbersome and laborious.

In parallel, recent advancements in Large Multimodal Models (LMMs)~\cite{wang2024qwen2, li2024llava-ov, li2024llava} have spurred a growing trend toward leveraging LMMs for a diverse range of vision-language tasks. For instance, LISA~\cite{lai2024lisa} uses LMMs for segmentation tasks, while LLaVA~\cite{liu2024visual} applies them to general Visual Question Answering (VQA). The success of LMMs in these domains can be attributed to two key factors: (i) the use of language as an interface between machines and humans, which naturally facilitates interaction across multiple tasks; and (ii) the training of Large Language Models (LLMs) on extensive text corpora, which has substantially improved their comprehension of natural language and, crucially, enriched their understanding of real-world knowledge. In this paper, we aim to extend this success by addressing a broad range of retrieval and reranking challenges within the scope of LMMs. This approach enables the unification of all retrieval tasks under the same formulation and allows for extrapolation to unseen retrieval tasks without the need for additional training.

We introduce a framework that enhances LMMs with universal retrieval and reranking capabilities, by simply inserting lightweight LoRA modules into LMMs, termed as \textbf{LamRA}. For quick retrieval, we adopt a two-stage training strategy to progressively improve the model's retrieval performance.
Specifically, the first stage involves text-only pre-training,
to enable the LMM to output embeddings in response to summarization prompts.
The second stage involves instruction-tuning, 
where we fine-tune the model on diverse datasets across various retrieval tasks. 
For reranking, we train an additional LoRA module on the LMM, 
to handle multiple images or lengthy texts. 
Using hard sample mining from the quick retrieval model, we conduct joint training for both pointwise and listwise reranking, enabling flexible support for single or multiple candidate inputs, further boosting the overall performance significantly.

% To answer the above question, this paper introduces \textbf{LamRA}, a comprehensive framework designed to equip LMM with advanced retrieval and reranking capabilities. Specifically, to enable LMM to perform effective retrieval, we propose a two-stage training strategy aimed at progressively refining their retrieval performance. 
% The first stage is pre-training, in which we strengthen the LMM's ability to extract single-modal embeddings.
% \weidi{unclear what do you mean by `extract single-modal embeddings'}
% The second stage involves instruction tuning, where the model is fine-tuned using a diverse set of retrieval datasets. This step aims to further improve the model's adaptability to a range of retrieval tasks. To further equip LMMs with reranking capabilities and leverage their ability to process flexible inputs, we employ hard sample mining on the trained retrieval model, enabling joint training for both pointwise and listwise reranking. The resulting reranker model supports inputs of single or multiple candidates to further optimize overall retrieval performance.

To validate the effectiveness of our method, we conduct extensive experiments across more than 10 distinct retrieval scenarios, including text-to-image, text-image-to-image, and text-image-to-text-image retrieval, {\em etc}. 
The robust performance indicates that our method can handle both simple and complex retrieval tasks effectively. Furthermore, we examine the model's task-level generalization by training on a subset of retrieval tasks and evaluating on previously unseen tasks. Remarkably, our method demonstrates strong generalization abilities, effectively performing on held-out tasks without prior exposure, underscoring its potential for transferability to unseen retrieval challenges.

% 三个 contribution 
To summarise, in this paper, we make the following contribution:
(i) We introduce \textbf{LamRA}, a versatile framework designed to empower LMMs with sophisticated retrieval and reranking capabilities. (ii) To enable efficient retrieval, we propose LamRA-Ret, which adopts a two-stage training strategy comprising language-only pre-training and multimodal instruction tuning to progressively enhance the model's retrieval capability. (iii) For reranking, we present LamRA-Rank, which supports both pointwise and listwise reranking to further boost the retrieval performance. (iv) Through extensive experiments, we demonstrate the effectiveness of our method in both supervised and zero-shot settings, achieving robust performance even on previously unseen retrieval tasks. Consequently, our method performs on par with or significantly surpasses existing state-of-the-art (SOTA) models across more than ten retrieval tasks.
\vspace{-0.5em}
% \weidi{the first two contribution are repeating, change to 
% (i) light-weight training to adapt LMMs for universal retrieval, 
% (ii) LMMs for quick retrieval with two-stage training;
% (iii) LMMs for both pointwise, listwise re-ranking with LMMs;
% (iv) experiments.}

\section{Related Work}
\noindent \textbf{Multimodal Information Retrieval.} 
Traditional multimodal information retrieval has often been limited to cross-modal retrieval settings, with evaluation benchmarks typically restricted to datasets like MSCOCO~\cite{lin2014microsoft} and Flickr30K~\cite{plummer2015flickr30k}. As the field progresses, more complex retrieval tasks have emerged, including composed image retrieval~\cite{liu2021image, wu2021fashion, baldrati2023zero, vaze2023genecis}, long-text-to-image retrieval~\cite{zhang2024long}, and image/question-to-multimodal document retrieval~\cite{chen2023can, hu2023open}, among others. As discussed in ~\cite{chen2023can, mensink2023encyclopedic, zhang2024long}, these tasks present substantial challenges for VLMs. Current approaches~\cite{miech2021thinking, brown2020smooth, xusentence, baldrati2022conditioned} often address this by fine-tuning models on individual tasks, yet developing universal multimodal embeddings capable of handling these various tasks simultaneously remains a significant challenge. 
% \weidi{add more reference. check SmoothAP, and more recent papers,  slow-fast for retrieval ?}

% \weidi{merge the following work into this related work section.}
% Inspired by PromptEOL~\cite{jia2021scaling} and E5-V~\cite{jiang2024e5}, which demonstrate that incorporating an Explicit One-word Limitation (EOL) into prompts to get the embedding from the generative models,

% \weidi{this title is not relevant to what you discuss in the following. It's unclear to me what this section is about, CLIP, ALIGN are trained with large-scale contrastive learning, then go to universal multimodal embeddings. }

\vspace{3pt}\noindent \textbf{Multimodal Representation Learning.} The challenge of developing robust multimodal representation remains a foundational question in multimodal learning. Pioneering models such as CLIP~\cite{radford2021learning} and ALIGN~\cite{jia2021scaling} have explored this by employing a dual-encoder architecture to learn effective representations through contrastive learning~\cite{oord2018representation} on large-scale image-text pairs. However, given that dual-encoder networks like CLIP aim to enhance the alignment between the image and text, they encounter challenges when dealing with interleaved image-text inputs. Moreover, the text encoder learned in this way shows a limited understanding of complex text~\cite{hsieh2024sugarcrepe}, suggesting the traditional dual-encoder models may require further refinement.

Numerous studies have pursued further exploration to achieve more universal multimodal representation. For instance,  UniIR~\cite{wei2023uniir} introduces a benchmark comprising eight distinct retrieval tasks, demonstrating that CLIP, when trained on this benchmark, achieves enhanced generality and adaptability across various retrieval tasks. E5-V~\cite{jiang2024e5} leverages Large Multimodal Models (LMMs), using carefully designed prompts to map images and text into a shared language hidden space. Through fine-tuning on text pairs~\cite{gao2021simcse}, E5-V demonstrates remarkable zero-shot retrieval capabilities. However, these methods demonstrate limited effectiveness on certain complex retrieval tasks. In this paper, we explore the potential of LMM as a universal retriever and reranker to address these challenges. 

% \vspace{-2em}

\begin{figure*}[ht]
  \centering
  \includegraphics[width=\textwidth]{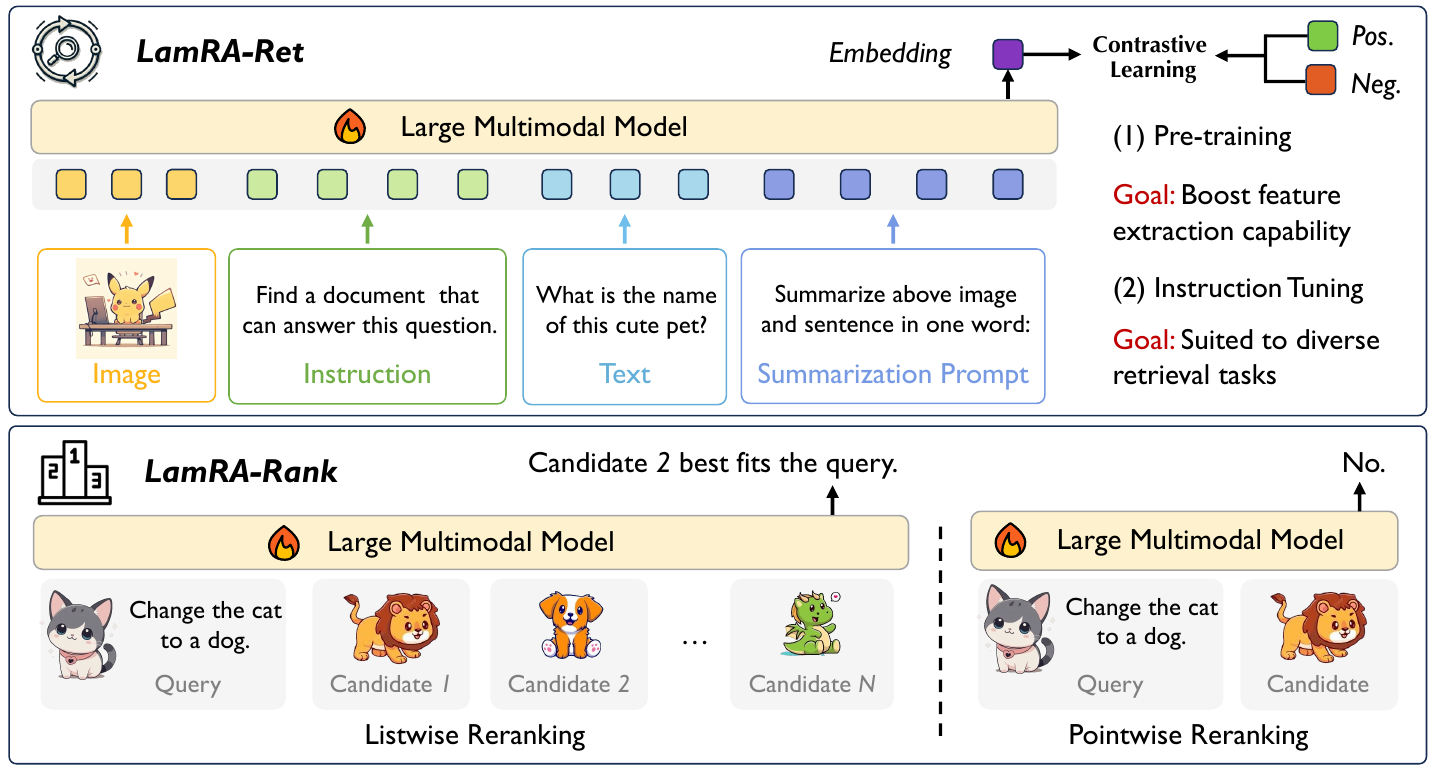} \\
  \vspace{-6pt}
  \caption{\textbf{Overview of the proposed LamRA framework.} LamRA consists of two components: LamRA-Ret and LamRA-Rank. The top section illustrates LamRA-Ret, encompassing both the pre-training and instruction-tuning stages, where contrastive learning is employed to enhance the retrieval capability of LMMs. The pre-training stage aims to improve the feature extraction capabilities through text-to-text retrieval, while the instruction tuning stage adapts the LMMs to various retrieval tasks by fine-tuning on diverse tasks with task-specific instructions. The bottom section depicts the joint training process of LamRA-Rank, which integrates both pointwise and listwise reranking.}
  \vspace{-1em}
 \label{fig:arch}
\end{figure*}

\noindent {\bf Large Multimodal Model.} 
With recent advancements in Large Language Models (LLMs)~\cite{brown2020language, jiang2023mistral, dubey2024llama}, recent research have focused on Large Multimodal Models (LMMs) to align visual and textual modalities through visual instruction tuning.
In particular, 
there is a growing trend toward using LMMs to tackle a wide range of vision-language tasks. For instance, LISA~\cite{lai2024lisa} applies LMMs to segmentation tasks, DetGPT~\cite{pi2023detgpt} utilizes LMMs for object detection, and VisionLLM~\cite{wang2024visionllm} addresses various vision-centric tasks using LMMs. Nevertheless, relatively few studies have explored the potential of LMMs in universal retrieval tasks, which is our focus in this paper, where we propose a framework LamRA, 
that adapts LMMs with light-weight LoRAs for advanced retrieval and reranking, significantly enhancing their performance in various retrieval tasks. We notice a few concurrent studies~\cite{jiang2024vlm2vec, lin2024mm} also explore the use of LMMs for universal retrieval, and we hope that further research will continue to advance this field.

%Although methods such as Fromage~\cite{koh2023grounding} have demonstrated strong performance in retrieval tasks within multi-round dialogues, MATE~\cite{jang2024mate} has aligned the CLIP image encoder and LLMs to enhance performance in image-to-long-text retrieval tasks, these studies remain largely confined to specific scenarios. 

%To unlock the potential of LMMs as universal retrievers and rerankers, 

%\kun{Due to the powerful VQA capabilities demonstrated by LMMs~\cite{liu2024visual, li2024llava-ov, li2024llava}, 

\section{Method}

This section starts by formulating our considered problem in Section~\ref{sec:problem_formulation}; then we elaborate on the architecture and feature extraction process in Section~\ref{sec:feature_extraction}; and describe the training details to equip the LMMs with retrieval capability in Section~\ref{sec:training_for_retrieval}; then, we present the approach to endow the LMMs with universal reranking abilities in Section~\ref{sec:reranking}, aiming to further boost retrieval performance; lastly, we detail the inference pipeline of our method in Section~\ref{sec:inference}.

\subsection{Problem Formulation} 
\label{sec:problem_formulation}

We address the challenge of universal retrieval and reranking with a generative model. Specifically, for one specific query~($q$), which can be an image, text, or an interleaved image-text format, and a retrieval set of $N$ candidates $\Omega = \{c_1, c_2, \cdots, c_N\}$ where each $c_i$ comprises images, text, or interleaved formats, we first extract the embeddings for both the query and all candidates with LMMs, and rank all candidates in $\Omega$ based on the relevance, 
{\em i.e.}, cosine similarity between the query and candidate embeddings. 
This initial retrieval process yields the top-$K$ candidates, represented as $\mathcal{C}_1 = \Phi_{\text{ret}}(q, \Omega)$. Subsequently, we refine this subset through a reranking process to produce the final ranked output, denoted as $\mathcal{C}_2 = \Phi_{\text{rerank}}(q, \mathcal{C}_1)$. Here, $\mathcal{C}_2$ represents the final reordered set of candidates.

In the following section, we provide details on extracting features for queries and candidates of arbitrary format.

% \weidi{need to provide more details, 
% how images and texts are encoded, and get the embedding, 
% then what is the task, i.e., rank the cases in a pool by relevance score, .... 
% what does reranking looks like,....}
% \weidi{need to provide a formulation for this universal retrieval problem, for example, assume a set of images $\mathcal{I} = \{I_1, I_2, \dots, I_n\}$, ...}

\subsection{Architecture \& Feature Extraction}
\label{sec:feature_extraction}

Generally speaking, the architecture of Large Multimodal Models~(LMMs) typically consists of three main components: a vision encoder, a vision projector, and a language model. To compute the embedding with a generative model, 
we adopt a similar approach as ~\cite{jiang2023scaling, jiang2024e5},
using an Explicit One-word Limitation (EOL). 
Specifically, we use the following prompts: 
(i) for image-only input, we use: \texttt{<image> Summarize above image in one word: <emb>}; 
(ii) for text-only input, we use: \texttt{<text> Summarize above sentence in one word: <emb>};
(iii) for mixed image-text input, we utilize: \texttt{<image\textsubscript{1}><text\textsubscript{1}>...<image\textsubscript{i}><text\textsubscript{j}> Summarize above image and sentence in one word: <emb>.} 
where \texttt{<image>} and \texttt{<text>} denote placeholders for the input image and sentence, respectively. For any combination of images and text inputs, the LMM uses the vision encoder and vision projector to map images into language space, which are then processed alongside the text by the large language model. 
We use the last hidden state immediately preceding the \texttt{<emb>} token as the representation of the input.

\subsection{Training for Retrieval} 
\label{sec:training_for_retrieval}

\vspace{2pt}\noindent 
In this section, we describe a two-stage training scheme, to advance the retrieval capabilities of LMMs. The first stage involves language-only pre-training, which facilitates the generative model to output improved embeddings. 
The second stage, instruction tuning, further adapts the LMMs to a wide range of retrieval tasks.

\vspace{2pt}
\noindent \textbf{Stage-I: Adapting LMMs for Retrieval Tasks.} 
LMMs were trained for generative tasks, {\em e.g.}, next-token prediction, which limits their ability for retrieval, as evidenced by the performance metrics in Table~\ref{tab:mbeir}, {\em i.e.}, directly employing Qwen2-VL-7B~\cite{wang2024qwen2} for retrieval tasks results in inferior performance. To address this issue, we start by adapting LMMs for text-to-text retrieval by training LoRA modules on the Natural Language Inference (NLI) dataset~\cite{gao2021simcse}, consisting pairs of text samples that are ideal to enhance LMMs for retrieval. For further discussion on the selection of pre-training datasets, please refer to Appendix~\ref{sec:supp_pretraining_data_selection}. 

\vspace{2pt}
\noindent \textbf{Stage-II: Instruction Tuning for Universal Retrieval.} 
To further enhance the retrieval capabilities of the LMMs across various multimodal retrieval tasks, we train on M-BEIR, which comprises 8 distinct retrieval tasks with 10 different datasets, as curated by~\cite{wei2023uniir}, 
%Unlike the pre-training stage, the instruction-tuning here extends training across eight different tasks, 
including image-to-image retrieval, composed image retrieval, and image/question-to-multimodal-document retrieval, among others. 
For different retrieval tasks, we incorporate task-specific instructions tailored for each retrieval type. For example, to complete an image-to-image retrieval task, the instruction can be ``\textit{retrieval a similar image.}" For further details regarding the instructions and M-BEIR datasets, please refer to Appendix~\ref{sec:supp_m_beir}.

\vspace{2pt}
\noindent \textbf{Training Objective.} 
We employ contrastive learning with the InfoNCE loss~\cite{oord2018representation} for both language-only pre-training and instruction tuning stages. 
Specifically, given a batch size of $B$, the embeddings of $n$-th query $q_n$ should be positioned close to the embeddings of its positive target $c_n$ and far away from other negative instances, formulated as: 
\begin{equation*}
\mathcal{L}_{\text{ret}}=-\frac{1}{B} \sum_{n=1}^B \log \left[\frac{\exp \left[ \kappa\left(\texttt{LMM}(q_n), \texttt{LMM}(c_n)\right) / \tau \right]}{\sum_{m=1}^B \exp \left[ \kappa\left(\texttt{LMM}(q_n), \texttt{LMM}(c_m)\right) / \tau \right]}\right]
\end{equation*}
% \weidi{the introduced notation has never been used here, for example, $\Phi_{\text{ret}}$ introduced in Sec.~3.1. or change your notations in the formulation accordingly.}

\noindent where $\tau$ refers to the temperature parameter, and $\kappa(\cdot,\cdot)$ denotes the cosine similarity. Additionally, $\texttt{LMM}(\cdot)$ represents the feature extraction process with LMMs, as mentioned in Section~\ref{sec:feature_extraction}. 
The model after both pre-training and instruction tuning is referred as \textbf{LamRA-Ret}.

% \weidi{I would suggest to add a paragraph on `Training'. 
% Unify the Stage-I, and Stage-II into one training scheme.}

\subsection{Training for Reranking}
\label{sec:reranking}

Here, we exploit the flexibility of LMMs to process multiple images or long texts and train a lightweight LoRA module to adapt them for reranking. 
As a result, we propose \textbf{LamRA-Rank}, 
which supports inputs of single or multiple candidates from the outputs of LamRA-Ret, to further improve the retrieval performance.

\vspace{2pt}
\noindent \textbf{Collecting Training Data for Reranking.} 
Here, we employ the LamRA-Ret model trained in Section~\ref{sec:training_for_retrieval} as the initial retriever, and train the reranking model on its top 100 retrieved candidates and use them as hard negatives. This approach enables LamRA-Rank to acquire a more nuanced understanding and improve its ranking performance.

\vspace{2pt}
\noindent \textbf{Joint Training for Pointwise and Listwise Reranking.} 
Given that the LMMs can accept flexible inputs, we explore joint training for both pointwise and listwise reranking.

In the case of pointwise reranking, for any query ($q$), 
we randomly select a negative candidate ($c_{\text{neg}}$) from the top 100 candidates and instruct the LamRA-Rank to output \texttt{YES} for the ground truth ($c_{\text{pos}}$) and \texttt{NO} for the negative candidate. Cross-entropy loss is employed as the loss function, expressed as $\mathcal{L}_{\text{point}} = \mathcal{L}_{\text{ce}}(\texttt{YES}, \texttt{Reranker}(q, c_{\text{pos}})) + \mathcal{L}_{\text{ce}}(\texttt{NO}, \texttt{Reranker}(q, c_{\text{neg}})),$ where $\texttt{Reranker}(\cdot,\cdot)$ represents the autoregressive output process of LamRA-Rank.
% \weidi{don't understand the notations $q$, $t$, $\texttt{Reranker}$ they have never been explained.}

In contrast, for listwise reranking, we randomly select $M$ negative samples $(c_1, c_2, \cdots, c_M)$ from the top 100 candidates, where $M$ is a randomly determined integer between 2 and 5. We then randomly insert the ground truth $(c_{\text{pos}})$ at any position and prompt the LamRA-Rank to directly output the position number of the ground truth. This process can be formulated as $\mathcal{L}_{\text{list}} = \mathcal{L}_{\text{ce}}(\texttt{GT-POSITION}, \texttt{Reranker}(q, c_{\text{pos}}, c_1, c_2, \cdots, c_M)).$

% \weidi{same problem as Sec.~3.3, you need to use your introduced notation, for example, $\Phi_{\text{rerank}}$ introduced in Sec.~3.1. or change your notations in the formulation accordingly.}

% \weidi{Same problem as before, there should be a place to formally introduce reranking with LMMs.}

The final loss is a weighted sum: $\mathcal{L}_{\text{rank}} = \mathcal{L}_{\text{point}} + \mathcal{L}_{\text{list}}$. 

\subsection{Inference Pipeline}
\label{sec:inference}

\noindent At the inference stage, we employ LamRA-Ret for the initial retrieval phase~($\Phi_{\text{ret}}$) and LamRA-Rank for the subsequent reranking~($\Phi_{\text{rank}}$). 
Given a query ($q$) and a candidate set ($\Omega$), LamRA-Ret is first utilized to compute an embedding similarity score ($S_{\text{ret}}$) for each candidate based on cosine similarity. Subsequently, the candidates are sorted to obtain the top-$K$ candidate set ($\mathcal{C}_1$). For the reranking process, we offer two options: pointwise reranking and listwise reranking. In the case of pointwise reranking,  the query and each of the top-$K$ candidates are sequentially input into the LMM, performing $K$ inference operations and assigning a reranking score ($S_{\text{rank}}$) to each candidate. This reranking score comes from the probability of the LMMs outputting \texttt{YES} for each candidate. Conversely, for listwise reranking, the query and the top-$K$ candidates are simultaneously fed into LamRA-Rank, which directly output the serial number of the most relevant candidate. More discussion on these two methods can be found in Section~\ref{sec:ablate}. In the final step, the embedding similarity score from LamRA-Ret and the reranking score from LamRA-Rank are aggregated into a weighted sum to get the final score: $S = \alpha \times S_{\text{ret}} + (1 - \alpha) \times S_{\text{rank}}$, where $\alpha$ is a weight hyperparameter. This final score is then used to generate the reordered candidate set ($\mathcal{C}_2$). We refer to the combined framework of LamRA-Ret and LamRA-Rank as \textbf{LamRA}.
\vspace{2pt}

\section{Experiments}
\subsection{Experimental Setup}
\label{experimental_setup}

\vspace{3pt}\noindent \textbf{Datasets and Metrics.} 
We utilize the NLI dataset~\cite{gao2021simcse} for pre-training and the M-BEIR~\cite{wei2023uniir} dataset for instruction tuning. 
The M-BEIR dataset encompasses eight distinct retrieval tasks across 10 different retrieval datasets, comprising a total of 1.1M training samples. As shown in Table~\ref{tab:dataset}, to evaluate the versatility of LamRA across various retrieval tasks, we conduct assessments on the M-BEIR test set. Furthermore, we investigate LamRA's generalization ability on other previously unseen datasets, including ShareGPT4V~\cite{zhang2024long}, Urban-1K~\cite{zhang2024long}, CIRCO~\cite{baldrati2023zero}, and Visual Dialog~\cite{das2017visual}, among others. 
We adhere to the standard evaluation metrics established for each dataset. For the retrieval tasks, we primarily utilize Recall@K as the evaluation metric, and for the image-text matching task, we employ Accuracy as the evaluation metric.

\vspace{2pt}
\noindent \textbf{Experiment Settings \& Baselines}. 
We establish three distinct experiment settings: 
(i) To validate the versatility of our method across a range of retrieval tasks, we train it on all 8 tasks in the M-BEIR benchmark and evaluate its performance on the test sets. For the baseline in this setting, we follow UniIR~\cite{wei2023uniir}, comparing with several strong VLMs, such as CLIP, SigLIP, and BLIP2, to assess their zero-shot performance. 
In the supervised setting, we compare our method with fully fine-tuned versions of UniIR-BLIP and UniIR-CLIP. 
(ii) To evaluate the generalization ability on previously \textbf{unseen retrieval datasets}, we perform zero-shot experiments on 10 datasets not encountered during training. In this case, the baseline includes a selection of universal retrievers, such as E5-V, MagicLens, and EVA-CLIP-18B. 
(iii) To investigate the generalization capacity on \textbf{unseen retrieval tasks}, we intentionally exclude data from three retrieval tasks: image-to-image retrieval, text-image-to-text retrieval, and text-image-to-text-image retrieval. Training is then conducted on the remaining five tasks with the evaluation of these excluded tasks.

\begin{table}[t]
\centering
\tiny  % 调整字体大小为更小的脚注级别
\resizebox{.47\textwidth}{!}{  % 控制表格的整体宽度
\setlength{\tabcolsep}{1mm}{
  \begin{tabular}{lccc}
    \toprule
    \textbf{Benchmark} & \textbf{Zero-shot} & \textbf{\# Queries} & \textbf{\# Candidates} \\
    \midrule
    M-BEIR~\cite{wei2023uniir} & \ding{56} & 190K & 5.6M\\
    ShareGPT4V~\cite{zhang2024long} & \ding{52} & 1K & 1K\\
    Urban-1K~\cite{zhang2024long} & \ding{52} & 1K & 1K\\
    Flickr30K~\cite{plummer2015flickr30k} & \ding{52} & 1K & 5K \\
    CIRCO~\cite{baldrati2023zero} & \ding{52} & 800 & 120K\\
    GeneCIS~\cite{vaze2023genecis} & \ding{52} & 8K & 10 $\sim$ 15\\
    Visual Storytelling~\cite{huang2016visual} & \ding{52} & 5K & 8K\\
    Visual Dialog~\cite{das2017visual} & \ding{52} & 2K & 2K\\
    Multi-round FashionIQ~\cite{yuan2021conversational} & \ding{52} & 2.4K & 6.2K\\
    CC-Neg~\cite{singh2024learn} & \ding{52} & 40K & 2 \\
    Sugar-Crepe~\cite{hsieh2024sugarcrepe} & \ding{52} & 7.5K & 2 \\
    \bottomrule
  \end{tabular}
}
}
\caption{\textbf{Summary of the evaluation benchmarks.} \# Queries represents the number of test queries, and \# Candidates denotes the number of test candidates per query.
}
\vspace{-2em}
\label{tab:dataset}
\end{table}

\begin{table*}[t]
% \small
% \setlength{\tabcolsep}{0.1pt}
\centering
\resizebox{\linewidth}{!}{
\begin{tabular}{lc@{\hspace{0.1cm}}c@{\hspace{0.1cm}}c@{\hspace{0.1cm}}c@{\hspace{0.1cm}}c@{\hspace{0.1cm}}c@{\hspace{0.1cm}}c@{\hspace{0.1cm}}c@{\hspace{0.1cm}}c@{\hspace{0.1cm}}c@{\hspace{0.1cm}}c@{\hspace{0.1cm}}c@{\hspace{0.1cm}}c@{\hspace{0.1cm}}c@{\hspace{0.1cm}}c@{\hspace{0.1cm}}c@{\hspace{0.1cm}}c@{\hspace{0.1cm}}}
\toprule
 & \multicolumn{3}{c}{{$q^t \to c^i$}} & {$q^t \to c^t$} & \multicolumn{2}{c}{{$q^t \to (c^i, c^t)$}} & \multicolumn{3}{c}{{$q^i \to c^t$}} & {$q^i \to c^i$} & \multicolumn{2}{c}{{$(q^i, q^t) \to c^t$}} & \multicolumn{2}{c}{{$(q^i, q^t) \to c^i$}} & \multicolumn{2}{c}{{$(q^i, q^t) \to (c^i, c^t)$}} & \\
 \cmidrule(r){2-4} \cmidrule(r){5-5}  \cmidrule(r){6-7} \cmidrule(r){8-10} \cmidrule(r){11-11} \cmidrule(r){12-13} \cmidrule(r){14-15} \cmidrule(r){16-17} 
 Methods & VN  & COCO & F200K & WebQA & EDIS & WebQA & VN & COCO & F200K & NIGHTS & OVEN & InfoS & FIQ & CIRR & OVEN & InfoS & Avg. \\
\cmidrule(r){2-4} \cmidrule(r){5-5}  \cmidrule(r){6-7} \cmidrule(r){8-10} \cmidrule(r){11-11} \cmidrule(r){12-13} \cmidrule(r){14-15} \cmidrule(r){16-17} 
& R@5 & R@5 & R@10 & R@5 & R@5 & R@5 & R@5 & R@5 & R@10 & R@5 & R@5 & R@5 & R@10 & R@5 & R@5 & R@5 & \\
\midrule
\multicolumn{18}{c}{\textit{Zero-shot}} \\
\midrule
CLIP-L~\cite{radford2021learning} & \textbf{{\color{blue}43.3}} & 61.1 & 6.6 & 36.2 & 43.3 & 45.1 & 41.3 & 79.0  & 7.7 & 26.1 & 24.2 & 20.5 & 7.0 & 13.2 & 38.8 & 26.4 & 32.5    \\
SigLIP~\cite{zhai2023sigmoid} & 30.1 & 75.7 & \textbf{{\color{red}36.5}} & 39.8 & 27.0 & 43.5 & 30.8 & 88.2  & \textbf{{\color{blue}34.2}} & 28.9 & 29.7 & 25.1 & 14.4 & 22.7 & 41.7 & 27.4 & 37.2  \\
BLIP~\cite{li2022blip} & 16.4 & 74.4 & 15.9 & 44.9 & 26.8 & 20.3 & 17.2 & 83.2  & 19.9 & 27.4 & 16.1 & 10.2 & 2.3 & 10.6 & 27.4 & 16.6 & 26.8  \\
BLIP2~\cite{li2023blip} & 16.7 & 63.8 & 14.0 & 38.6 & 26.9 & 24.5 & 15.0 & 80.0  & 14.2 & 25.4 & 12.2 & 5.5 & 4.4 & 11.8 & 27.3 & 15.8 & 24.8  \\
Qwen2-VL-7B~\cite{wang2024qwen2} & 9.3 & 55.1 & 5.0 & 42.0 & 26.2 & 9.4 & 5.4 & 46.6 & 4.0 & 21.3 & 21.4 & 22.5 & 4.3 & 16.3 & 43.6 & 36.2 & 23.0 \\
\midrule
\multicolumn{18}{c}{\textit{Supervised - Dual Encoder}} \\
\midrule
$\text{UniIR-BLIP}_{\text{FF}}$~\cite{wei2023uniir} & 23.4 & 79.7 & 26.1 & 80.0 & 50.9 & 79.8 & 22.8 & 89.9 & 28.9 & \textbf{{\color{blue}33.0}} & 41.0 & 22.4 & 29.2 & 52.2 & 55.8 & 33.0 & 46.8  \\
$\text{UniIR-CLIP}_{\text{SF}}$~\cite{wei2023uniir} & 42.6 & 81.1 & 18.0 & 84.7 & 59.4 & 78.7 & \textbf{{\color{blue}43.1}} & \textbf{{\color{blue}92.3}} & 18.3 & 32.0 & 45.5 & 27.9 & 24.4 & 44.6 & 67.6 & 48.9 & 50.6  \\
\midrule
\multicolumn{18}{c}{\textit{Supervised - LMMs}} \\
\midrule
% \addlinespace[0.2em]
% \arrayrulecolor{gray}\hdashline
% \specialrule{0em}{0.5pt}{0.5pt} 
% \largehrulespace 
% LamRA-Ret w/o Pre. & 35.6 & 76.3 & 22.0 & 86.2 & 60.7 & 79.7 & 36.8 & 89.6 & 27.8 & 31.2 & 53.0 & 49.8 & 31.4 & 44.8 & 74.5 & 57.4 & 53.6 \\
% LamRA-Ret w/o Inst. & 16.9 & 73.5 & 10.6 & 80.1 & 45.0 & 60.3 & 14.3 & 81.7 & 12.2 & 28.6 & 24.1 & 31.7 & 7.5 & 28.7 & 36.7 & 27.7 & 36.2\\   
LamRA-Ret & 41.6 & \textbf{{\color{blue}81.5}} & 28.7 & \textbf{{\color{blue}86.0}} & \textbf{{\color{blue}62.6}} & \textbf{{\color{blue}81.2}} & 39.6 & 90.6 & 30.4 & 32.1 & \textbf{{\color{blue}54.1}} & \textbf{{\color{blue}52.1}} & \textbf{{\color{blue}33.2}} & \textbf{{\color{blue}53.1}} & \textbf{{\color{blue}76.2}} & \textbf{{\color{blue}63.3}} & \textbf{{\color{blue}56.6}} \\
LamRA & \textbf{{\color{red}48.0}} & \textbf{{\color{red}85.2}} & \textbf{{\color{blue}32.9}} & \textbf{{\color{red}96.7}} & \textbf{{\color{red}75.8}} & \textbf{{\color{red}87.7}} & \textbf{{\color{red}48.6}} & \textbf{{\color{red}92.3}} & \textbf{{\color{red}36.1}} & \textbf{{\color{red}33.5}} & \textbf{{\color{red}59.2}} & \textbf{{\color{red}64.1}} & \textbf{{\color{red}37.8}} & \textbf{{\color{red}63.3}} & \textbf{{\color{red}79.2}} & \textbf{{\color{red}78.3}} & \textbf{{\color{red}63.7}} \\

\bottomrule
 \end{tabular}
}
\vspace{-5pt}
\caption{\textbf{Comparison with up-to-date state-of-the-arts on M-BEIR test set.} The first row indicates the retrieval task type: $q^t$ for text queries, $q^i$ for image queries, $c^t$ for text candidates, and $c^i$ for image candidates. Abbreviations used include VN for VisualNews, F200K for Fashion200K, InfoS for InfoSeek, and FIQ for FashionIQ. Evaluation standards follow UniIR, with FashionIQ and Fashion200K using Recall@10, while all other evaluations employ Recall@5. The best (resp. second-best) numbers are in red (resp. blue).} 
\label{tab:mbeir}
\vspace{-1em}
\end{table*}

\vspace{2pt}\noindent \textbf{Implementation Details.} 
Our framework is implemented in Pytorch and leverages the Qwen2-VL-7B~\cite{wang2024qwen2} by default. In the pretraining stage for retrieval, we conduct experiments on 8 A100 GPUs with a batch size of 576 and a learning rate of $4\times 10^{-5}$, training for two epochs. 
During the instruction tuning stage, we train on 16 A100 GPUs with a batch size of 960 and a learning rate of $1\times 10^{-4}$ for one epoch. 
For the rerank training stage, we train for one epoch on 16 A100 GPUs, employing a batch size of 64 and a learning rate of $4\times 10^{-5}$. Throughout all experiments, the parameters on the vision side remain fixed, while the LLM is fine-tuned using LoRA. 
LamRA-Rank adopts the pointwise reranking by default. 
During the M-BEIR evaluation, experiments are conducted in the local pool, with LamRA-Rank reranking the top-50 results. For experiments on unseen datasets, reranking is applied to the top-10 results. The weight hyperparameter $\alpha$ is set to a default value of 0.5. 

% \weidi{what do you mean by dynamic search ?}

\subsection{Experimental Results}
\label{sec:sota}

\noindent \textbf{Versatility Across Various Retrieval Tasks.} As demonstrated in Table~\ref{tab:mbeir}, our model is capable of handling retrieval tasks across various input formats. We can draw the following observations: (i) Our approach significantly outperforms the dual-encoder paradigm method, UniIR-CLIP, in both multimodal and pure text retrieval tasks. For instance, in text-image-to-text retrieval on the InfoSeek dataset, our method LamRA-Ret, achieves an improvement of 24.2 points over UniIR-CLIP in Recall@5. Similarly, in composed image retrieval on the CIRR dataset, our method exceeds UniIR-CLIP by 8.5 points in Recall@5. (ii) In simpler retrieval tasks, such as cross-modal retrieval, the performance of LamRA-Ret is either comparable to or slightly higher than that of UniIR-CLIP. (iii) Furthermore, LamRA~(with reranking) achieves additional performance gains across all retrieval tasks, with an average improvement of 7.1 points over 16 M-BEIR retrieval tasks. (iv) Notably, Qwen2-VL-7B achieves an average zero-shot score of only 23.0 on the M-BEIR benchmark, which increases to 63.7 after applying our framework. This highlights the effectiveness of our proposed LMMs framework in enhancing general retrieval and reranking performance.

\vspace{2pt}
\noindent \textbf{Advanced Generalization Capability on Unseen Dataset.} To validate the strong generalization capability of our method, we conduct extensive experiments across 10 unseen datasets. As illustrated in Table~\ref{tab:zero-shot}, we can make the following observations: (i) Our method demonstrates excellent generalization across various unseen datasets, with performance either significantly exceeding or comparable to other strong baseline methods. (ii) Notably, our method performs exceptionally well in scenarios where dual-encoder approaches tend to underperform. For instance, in the image-to-long-text retrieval task on the Urban-1K dataset, our LamRA-Ret achieves over a 10-point improvement compared to other methods. Similarly, in tasks such as dialog-to-image retrieval and multi-round composed image retrieval, our method also yields substantial performance gains. (iii) In Image-Text Matching tasks, LMMs-based methods outperform those using the dual-encoder paradigm. We attribute this improvement primarily to the LMM's exceptional ability to comprehend complex text.

\begin{table}[t]
% \scriptsize
% \setlength{\tabcolsep}{0.1pt}
\setlength{\tabcolsep}{1.5mm}
\centering
\resizebox{.47\textwidth}{!}{
\begin{tabular}{lcccccc}
\toprule
 & {$q^i \to c^i$} & \multicolumn{2}{c}{{$(q^i, q^t) \to c^t$}} & \multicolumn{2}{c}{{$(q^i, q^t) \to (c^i, c^t)$}}\\
 \cmidrule(r){2-2} \cmidrule(r){3-4}  \cmidrule(r){5-6}  
 Methods & NIGHTS & OVEN & InfoS & OVEN & InfoS & Avg. \\
& R@5 & R@5 & R@5 & R@5 & R@5 & \\
\midrule
\multicolumn{7}{l}{\color{gray}{\textit{Supervised}}} \\
\addlinespace[0.15em]
$\text{UniIR-BLIP}_{\text{FF}}$~\cite{wei2023uniir} & 33.0 & 41.0 & 22.4 & 55.8 & 33.0 & 37.0 \\
$\text{UniIR-CLIP}_{\text{SF}}$~\cite{wei2023uniir} & 32.0 & 45.5 & 27.9 & 67.6 & 48.9 & 44.4 \\
\midrule
\multicolumn{7}{l}{\color{gray}{\textit{Zero-shot}}} \\
\addlinespace[0.15em]
LamRA-Ret$^*$ & 27.2 & 44.7 & 44.0 & 62.8 & 49.5 & 45.6  \\
LamRA$^*$ & 29.2 & 46.9 & 54.2 & 65.1 & 59.1 & 50.9 \\

\bottomrule
 \end{tabular}
}
\caption{\textbf{Held-out task generalization experiments on M-BEIR.} $^*$ means that the training is carried out on the remaining five tasks, with no exposure to these three held-out tasks.} 
\label{tab:task_generalization}
\vspace{-1cm}
\end{table}

\begin{table*}[t]
% \small
% \setlength{\tabcolsep}{1.2pt}
\setlength{\tabcolsep}{1.5mm}
\centering
\resizebox{\linewidth}{!}{
\begin{tabular}{lccccccccccccc}
\toprule
 & \multicolumn{3}{c}{{$q^t \to c^i$}} & \multicolumn{3}{c}{{$q^i \to c^t$}} & \multicolumn{2}{c}{{$(q^i, q^t) \to c^i$}} & \multicolumn{1}{c}{{$q^{\text{dialog}} \to c^i$}} & 
\multicolumn{2}{c}{{$(q^i \oplus q^t) \to c^i$}} &\multicolumn{2}{c}{{ITM}}\\
\cmidrule(r){2-4} \cmidrule(r){5-7}  \cmidrule(r){8-9} \cmidrule(r){10-10}  \cmidrule(r){11-12} \cmidrule(r){13-14} 
Methods & Share4V  & Urban$^*$ & Flickr & Share4V  & Urban$^*$ & Flickr  & CIRCO$^*$ & GeneCIS$^*$ & VisD$^*$ & VIST & MT-FIQ$^*$ & CC-Neg & Sugar-Crepe$^*$ \\
\cmidrule(r){2-4} \cmidrule(r){5-7}  \cmidrule(r){8-9} \cmidrule(r){10-10}  \cmidrule(r){11-12} \cmidrule(r){13-14} 
& R@1 & R@1 & R@1 & R@1 & R@1 & R@1 & MAP@5 & R@1 & R@1 & R@1 & R@5 & Acc. & Acc. \\
\midrule
CLIP-L~\cite{radford2021learning} & 84.0 & 52.8 & 67.3 & 81.8 & 68.7 & 87.2 & 4.0 & 13.3 & 23.7 & 0.6 & 17.7 & 66.7 & 73.0 \\
Long-CLIP-L~\cite{zhang2024long} & \textbf{{\color{blue}95.6}} & 86.1 & 76.1 & \textbf{{\color{blue}95.8}} & 82.7 & 89.3 & 5.7 & 16.3 & 37.9 & 1.1 & 18.5 & 76.3 & 80.9 \\
UniIR-CLIP~\cite{wei2023uniir} & 85.8 & 75.0 & 78.7 & 84.1 & 78.4 & 94.2 & 12.5 & 16.8 & 26.8 & 0.6 & 39.4 & 79.9 & 80.3 \\
E5-V~\cite{jiang2024e5} & 86.7 & 84.0 & 79.5 & 84.0 & 82.4 & 88.2 & 24.8 & 18.5 & 54.6 & 10.0 & 19.2 & \textbf{{\color{blue}83.2}} & 84.7 \\
MagicLens-L~\cite{zhang2024magiclens} & 85.5 & 59.3 & 72.5 & 60.9 & 24.2 & 84.6 & 29.6 & 16.3 & 28.0 & 3.3 & 22.6 & 62.7 & 75.9 \\
EVA-CLIP-8B~\cite{sun2024eva} & 91.2 & 77.8 & 80.8 & 93.1 & 80.4 & 95.6 & 6.0 & 13.1 & 23.2 & 1.2 & 22.1 & 59.4 & 81.7\\
EVA-CLIP-18B~\cite{sun2024eva} & 92.1 & 81.7 & \textbf{{\color{blue}83.3}} & 94.0 & 83.3 & \textbf{{\color{blue}96.7}} & 6.1 & 13.6 & 24.7 & 1.0 & 21.9 & 63.8 & 83.1\\
% \addlinespace[0.2em]
% \hdashline
% \addlinespace[0.2em]
\midrule
% \specialrule{0em}{0.5pt}{0.5pt} 
% \largehrulespace 
LamRA-Ret & 93.3 & \textbf{{\color{blue}95.1}} & 82.8 & 88.1  & \textbf{{\color{blue}94.3}} & 92.7 & \textbf{{\color{blue}33.2}} & \textbf{{\color{blue}18.9}}  & \textbf{{\color{blue}62.8}} & \textbf{{\color{blue}23.1}} & \textbf{{\color{blue}60.9}} & 79.6 & \textbf{{\color{blue}85.8}} \\
LamRA & \textbf{{\color{red}97.9}} & \textbf{{\color{red}98.8}} & \textbf{{\color{red}88.1}} & \textbf{{\color{red}96.5}} & \textbf{{\color{red}98.0}} & \textbf{{\color{red}97.6}} & \textbf{{\color{red}42.8}} & \textbf{{\color{red}24.8}} & \textbf{{\color{red}70.9}}  & \textbf{{\color{red}28.6}} & \textbf{{\color{red}63.9}}
& \textbf{{\color{red}85.9}} & \textbf{{\color{red}93.5}}\\

\bottomrule
 \end{tabular}
}
\vspace{-5pt}
\caption{\textbf{Comparison with up-to-date state-of-the-arts on unseen dataset.} The first row indicates the retrieval task type: $q^t$ for text queries, $q^i$ for image queries, $q^{\text{dialog}}$ for dialog queries, $(q^i \oplus q^t)$ for multi-interleaved image-text queries, $c^t$ for text candidates, $c^i$ for image candidates, and ITM for the Image-Text Matching task. Abbreviations used include Share4V for ShareGPT4V, Urban for Urban-1k, VIST for Visual Storytelling, VisD for Visual Dialog, and MT-FIQ for Multi-round FashionIQ. $*$ indicates that the images in these datasets are sourced from COCO or FashionIQ. However, due to significant differences in their captions or query formats, we still classify these datasets as unseen. We adhere to the standard evaluation metrics specific to each dataset. The best (resp. second-best) numbers are in red (resp. blue). For more details about these datasets, please refer to Appendix~\ref{sec:supp_unseen_datasets}.} 
\label{tab:zero-shot}
\vspace{-0.1cm}
\end{table*}

\vspace{2pt}
\noindent \textbf{Exceptional Generalization on Unseen Retrieval Task.} 
As outlined in Section~\ref{experimental_setup}, we train our model on five tasks and evaluate it on excluded tasks to assess its generalization on these previously unseen tasks. As demonstrated in Table~\ref{tab:task_generalization}, our approach shows impressive performance on unseen retrieval tasks. Notably, when compared to supervised methods such as UniIR-CLIP, our method achieves competitive or even superior zero-shot results in more complex tasks, such as text-image-to-text retrieval task on the InfoSeek benchmark, our method exceeds UniIR-CLIP by 26.3 points in Recall@5. This remarkable generalization ability suggests that our method can extrapolate to unseen retrieval tasks without additional training and may hold significant practical applicability across a broader range of scenarios. 

\subsection{Ablation Study \& Analysis}
\label{sec:ablate}

\noindent In this section, we conduct experiments to further investigate the effect of our proposal, for example, the two-stage training for LamRA-Ret, the scaling trends of LamRA, {\em etc}. 

% \weidi{consider to give an overview sentence, 
% `In this section, we conduct a series of ablation studies, to investigate (i), (ii), (iii).'}

\noindent \textbf{Effectiveness of Two-stage Training.} 
According to the results presented in Table~\ref{tab:ablation_two_stage}, both the pre-training and the instruction tuning stages significantly impact retrieval performance. Specifically, omitting the pre-training stage results in an average performance decline of three points on the M-BEIR benchmark. Similarly, the absence of the instruction tuning stage yields suboptimal results, as the model struggles with some complex multimodal retrieval tasks. Collectively, these stages together enable a progressive enhancement in the retrieval performance of the LMMs.

%Overall, the pre-training stage equips the LMMs with essential retrieval capability, whereas the instruction tuning stage further enhances its adaptability to various retrieval tasks.

\begin{table}[ht]
\centering
\setlength{\tabcolsep}{3.5mm}
\scriptsize  % 调整字体大小为更小的脚注级别
\resizebox{.45\textwidth}{!}{  % 控制表格的整体宽度
  \begin{tabular}{ccl}
    \toprule
    \textbf{Pre-training} & \textbf{Instruction tuning} & \textbf{Avg.}\\
    \midrule
    \ding{56} & \ding{56} & 23.0 \textcolor{gray}{(-33.6)}\\
     \ding{52} & \ding{56} & 36.2 \textcolor{gray}{(-20.4)}\\
     \ding{56} & \ding{52} & 53.6 \textcolor{gray}{(-3.0)} \\
     \ding{52} & \ding{52} & 56.6 \\
    \bottomrule
  \end{tabular}
}
\caption{\textbf{Effect of the two-stage training.} Avg. refers to the average recall performance across the M-BEIR test set.}
\label{tab:ablation_two_stage}
\vspace{-0.3cm}
\end{table}
\begin{table}[t]
\centering
\scriptsize
\resizebox{.45\textwidth}{!}{ 
  \begin{tabular}{cccc}
    \toprule
    \textbf{LMMs} & \textbf{LamRA-Ret} & \textbf{LamRA-Rank} & \textbf{Avg.} \\
    \midrule
     \multirow{2}{*}{Qwen2-VL-2B~\cite{wang2024qwen2}} & \ding{52} & \ding{56} & 51.6\\
      & \ding{52} & \ding{52} & 58.3\\
      \midrule
     \multirow{2}{*}{Qwen2-VL-7B~\cite{wang2024qwen2}} & \ding{52} & \ding{56} & 56.6\\
      & \ding{52} & \ding{52} & 63.7\\
    \bottomrule
  \end{tabular}
}
% }
\caption{\textbf{Scaling Trends of LamRA.} Experimental results on the M-BEIR test set across various model sizes.}
\label{tab:scaling_laws}
\end{table}

\vspace{2pt}
\noindent \textbf{Scaling Trends of LamRA.} 
As shown in Table~\ref{tab:scaling_laws}, to investigate the scaling effect associated with LamRA, we conduct experiments on LMMs of different sizes, specifically Qwen2-VL-2B and Qwen2-VL-7B. The results indicate that as the model size scales, performance improves correspondingly. 
Currently, due to the computational limitations, we are able to conduct experiments solely on the 2B and 7B models; however, our findings suggest that larger and more powerful LMMs potentially yield even superior results within our framework, thereby laying the groundwork for further exploration in this area.

\begin{table}[ht]
\centering
% \scriptsize
\setlength{\tabcolsep}{0.5mm}
\resizebox{.45\textwidth}{!}{  % 控制表格的整体宽度
  \begin{tabular}{cccccc}
    \toprule
    \multirow{2}{*}{\textbf{Task}} & \textbf{LamRA-Ret} & \multicolumn{2}{c}{\textbf{LamRA-Rank(P)}} & \multicolumn{2}{c}{\textbf{LamRA-Rank(L)}} \\
    \cmidrule(r){2-2} \cmidrule(r){3-4}  \cmidrule(r){5-6}
    & R@1 & R@1 & Time & R@1 & Time \\
    \midrule
      $q^t \to c^t$ & 58.2 & 75.9 & 0.020s & 75.9 & 0.010s\\

      $(q^i, q^t) \to c^i$ & 18.5 & 24.5 & 0.071s & 24.3 & 0.067s\\

      $(q^i, q^t) \to c^t$ & 30.1 & 37.3 & 0.047s & 36.6 & 0.017s\\

      $(q^i, q^t) \to (c^i, c^t)$ & 33.4 & 39.9 & 0.084s & 39.5 & 0.085s \\
    \bottomrule
  \end{tabular}
}
% }
\caption{\textbf{Comparison of Recall@1 performance and inference costs between pointwise (LamRA-Rank(P)) and listwise (LamRA-Rank(L)) reranking on M-BEIR.} Reranking is applied to the top-5 results. Time denotes the per-query inference time cost measured on eight A100 GPUs with a batch size of 32.}
\label{tab:point_list}
\vspace{-1.5em}
\end{table}

\vspace{2pt}
\noindent \textbf{Discussion on Pointwise and Listwise Rerank Methods.} 
As outlined in Section~\ref{sec:reranking}, LamRA-Rank is designed to perform both pointwise and listwise reranking. We investigate these two distinct reranking methods on the M-BEIR test set, with the results presented in Table~\ref{tab:point_list}. Both methods demonstrate a significant improvement in retrieval performance. Although pointwise reranking generally incurs a higher computational overhead than listwise reranking, the latter is constrained by the LLM's context length, which limits the number of candidates it can process. 
Consequently, each reranking method offers distinct advantages and can be flexibly applied in different contexts based on specific application requirements.

\begin{figure*}[t]
  \centering
  \includegraphics[width=\textwidth]{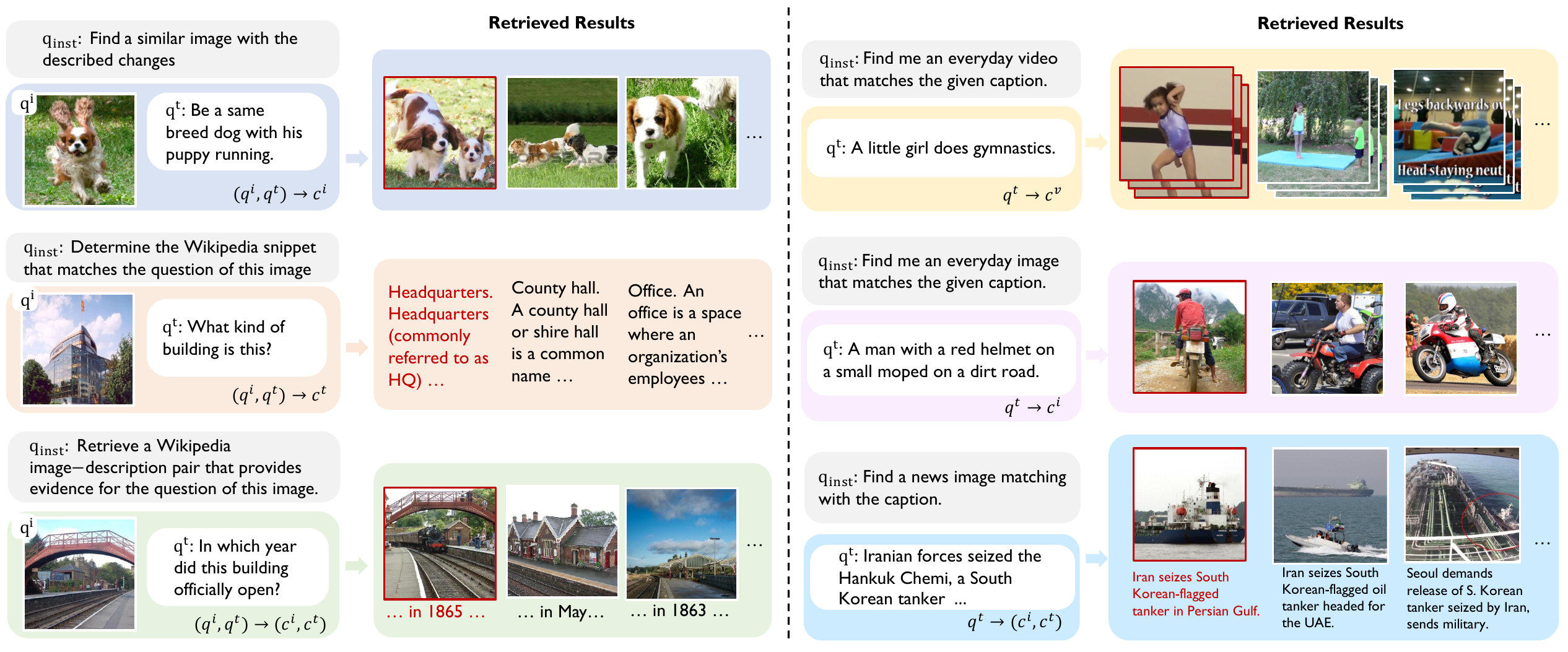} \\
  \vspace{-6pt}
  \caption{\textbf{Qualitative examples.} We show the results of our method across six different retrieval tasks, with the ground truth indicated by a red box and red text. Here, $q^t$ for text queries, $q^i$ for image queries, $c^t$ for text candidates, $c^i$ for image candidates, and $c^v$ for video candidates. For more qualitative examples, please refer to Appendix~\ref{sec:supp_qualitative}.}
    \vspace{-1em}
 \label{fig:qualitative}
\end{figure*}

\vspace{2pt}
\noindent \textbf{Extending to Video Retrieval.} 
As shown in Table~\ref{tab:zero_shot_video_ret}, we evaluate our method on the MSR-VTT~\cite{xu2016msr} and MSVD~\cite{chen2011collecting} datasets in a zero-shot setting for the text-to-video retrieval task. 
The results demonstrate that our approach exhibits competitive performance. For instance, on the MSR-VTT dataset, our method outperforms InternVideo~\cite{wang2022internvideo} by 4.7 points in Recall@1. 
Similarly, on the MSVD dataset, we achieve an improvement of 2.5 points over UMT-L~\cite{li2023unmasked} in Recall@1. Notably, our method has not been exposed to any video data during the fine-tuning stage, indicating that it preserves Qwen2-VL's inherent video understanding capability. Although our approach currently falls short of the performance achieved by the state-of-the-art InternVideo2~\cite{wang2024internvideo2}, we intend to address this gap by exploring the integration of video data in future research.
% \weidi{could this be in `unseen retrieval tasks '?}

\begin{table}[htbp]
\small
\centering
\resizebox{.47\textwidth}{!}{
\setlength\tabcolsep{1.5mm}
    \begin{tabular}{lcccccc}
        \toprule
        \multirow{2}{*}{\bf{Method}} & \multicolumn{3}{c}{\bf{MSR-VTT}} & \multicolumn{3}{c}{\bf{MSVD}} \\
        \cmidrule(r){2-4} \cmidrule(r){5-7}
         &  R@1 & R@5 & R@10 & R@1 & R@5 & R@10  \\
         \midrule
        InternVideo~\cite{wang2022internvideo} & 40.0 & 65.3 & 74.1 & 43.4 & 69.9 & 79.1 \\
        ViCLIP~\cite{wang2023internvid} & 42.4 & - & - & 49.1 & - & - \\
        UMT-L~\cite{li2023unmasked} & 42.6 & 64.4 & 73.1 & 49.9 & 77.7 & 85.3 \\
        InternVideo2$_{s2}$-6B~\cite{wang2024internvideo2} & 55.9 & 78.3 & 85.1 & 59.3 & 84.4 & 89.6\\
        \midrule
    LamRA & 44.7 & 68.6 & 78.6 & 52.4 & 79.8 & 87.0\\
      \bottomrule
      \end{tabular}}
      \caption{\textbf{Zero-shot text-to-video retrieval performance.}}
      \label{tab:zero_shot_video_ret}
      \vspace{-1em}
\end{table} 

% \noindent \textbf{Integrating Retrieval and QA in one LMM.} \kun{consider putting this part in the appendix.}According to the analysis in Section~\ref{sec:sota}, the LMM demonstrates significant potential to serve as a universal retriever, bolstered by its excellent generative capabilities. This raises an important question: Can the retrieval and generative capabilities be integrated within the same LMM? To address this inquiry, we conduct experiments on three Knowledge-based Visual Question Answering (KVQA) tasks. KVQA requires the retrieval of relevant documents before answering associated questions, aligning closely with the question we aim to investigate. Specifically, we train the retrieval and VQA tasks simultaneously during the training process. The experimental results, presented in Table X, indicate that the retrieval performance of our model on the three datasets surpasses the current SOTA. Furthermore, the accuracy of answering questions based on the retrieval documents is comparable to or exceeds the current SOTA. These findings demonstrate the feasibility of integrating retrieval and generative capabilities within a single LMM, which we consider a promising direction for future work.

\noindent \textbf{Discussion on Inference Cost.} One potential drawback of applying LMMs to retrieval and reranking tasks is the high inference costs. Nevertheless, various strategies can be employed in practical applications to mitigate this issue. For instance, pre-extracting features for the candidate pool in advance or deploying our method in scenarios with low query-per-second requirements can significantly reduce inference costs. Moreover, several recent studies~\cite{kwon2023efficient} focus on improving the inference speed of LMMs. We believe that leveraging LMMs for retrieval and reranking holds promising potential for future applications.

\subsection{Qualitative Results.} 

\noindent In Figure~\ref{fig:qualitative}, we present the qualitative results of our method across six retrieval tasks. In the top left panel, we demonstrate our method's effectiveness in handling the composed image retrieval task, where the query requires retrieving an image of a dog of the same breed as shown in the reference image, along with its puppy running. Our method effectively captures this complex intent, successfully retrieving the target image. In the third row of the first column, we showcase our method's performance on the text-image-to-text-image retrieval task, where it not only retrieves a relevant Wikipedia snippet for the query but also identifies an image similar to the query image. For more qualitative examples, please refer to Appendix~\ref{sec:supp_qualitative}.

% In the first two rows of the second column, we demonstrate our method's capability to perform both text-to-image and text-to-video retrieval tasks.

% \input{tables/longcaption}
% \input{tables/ccneg}
% \input{tables/circo}
% \input{tables/genecis}
% \input{tables/multi-round-fashion}
% \input{tables/rag_tables}
% \input{tables/visdial}
% \input{tables/sugar_crepe}
% \input{tables/video_retrieval}
% \input{tables/editworld}
% \input{others/qaego4d_sota}
% \input{tables/ablate}

\section{Conclusion}
\noindent In this paper, we have explored a novel paradigm for applying LMMs to address a broad range of retrieval tasks. By simply inserting lightweight LoRA modules, Our proposed universal framework, LamRA, endows LMMs with robust retrieval and reranking capabilities. Specifically, for efficient retrieval, we introduced LamRA-Ret, which adopts a two-stage training strategy to progressively enhance retrieval performance. For reranking, we presented LamRA-Rank, which supports both pointwise and listwise reranking to further optimize results. Comprehensive quantitative and qualitative comparisons across more than ten diverse retrieval tasks underscore the versatility and effectiveness of our method. We anticipate that our framework will provide valuable insights into the multimodal information retrieval field and inspire further research in this area.

{
    \small
    \bibliographystyle{ieeenat_fullname}
    \bibliography{main}

\begin{thebibliography}{60}
\providecommand{\natexlab}[1]{#1}
\providecommand{\url}[1]{\texttt{#1}}
\expandafter\ifx\csname urlstyle\endcsname\relax
  \providecommand{\doi}[1]{doi: #1}\else
  \providecommand{\doi}{doi: \begingroup \urlstyle{rm}\Url}\fi

\bibitem[Baldrati et~al.(2022)Baldrati, Bertini, Uricchio, and Del~Bimbo]{baldrati2022conditioned}
Alberto Baldrati, Marco Bertini, Tiberio Uricchio, and Alberto Del~Bimbo.
\newblock Conditioned and composed image retrieval combining and partially fine-tuning clip-based features.
\newblock In \emph{Proceedings of the IEEE Conference on Computer Vision and Pattern Recognition}, 2022.

\bibitem[Baldrati et~al.(2023)Baldrati, Agnolucci, Bertini, and Del~Bimbo]{baldrati2023zero}
Alberto Baldrati, Lorenzo Agnolucci, Marco Bertini, and Alberto Del~Bimbo.
\newblock Zero-shot composed image retrieval with textual inversion.
\newblock In \emph{Proceedings of the International Conference on Computer Vision}, 2023.

\bibitem[Brown et~al.(2020{\natexlab{a}})Brown, Xie, Kalogeiton, and Zisserman]{brown2020smooth}
Andrew Brown, Weidi Xie, Vicky Kalogeiton, and Andrew Zisserman.
\newblock Smooth-ap: Smoothing the path towards large-scale image retrieval.
\newblock In \emph{Proceedings of the European Conference on Computer Vision}, 2020{\natexlab{a}}.

\bibitem[Brown et~al.(2020{\natexlab{b}})Brown, Mann, Ryder, Subbiah, Kaplan, Dhariwal, Neelakantan, Shyam, Sastry, Askell, Agarwal, Herbert-Voss, Krueger, Henighan, Child, Ramesh, Ziegler, Wu, Winter, Hesse, Chen, Sigler, Litwin, Gray, Chess, Clark, Berner, McCandlish, Radford, Sutskever, and Amodei]{brown2020language}
Tom Brown, Benjamin Mann, Nick Ryder, Melanie Subbiah, Jared~D Kaplan, Prafulla Dhariwal, Arvind Neelakantan, Pranav Shyam, Girish Sastry, Amanda Askell, Sandhini Agarwal, Ariel Herbert-Voss, Gretchen Krueger, Tom Henighan, Rewon Child, Aditya Ramesh, Daniel Ziegler, Jeffrey Wu, Clemens Winter, Chris Hesse, Mark Chen, Eric Sigler, Mateusz Litwin, Scott Gray, Benjamin Chess, Jack Clark, Christopher Berner, Sam McCandlish, Alec Radford, Ilya Sutskever, and Dario Amodei.
\newblock Language models are few-shot learners.
\newblock In \emph{Advances in Neural Information Processing Systems}, 2020{\natexlab{b}}.

\bibitem[Chen and Dolan(2011)]{chen2011collecting}
David Chen and William~B Dolan.
\newblock Collecting highly parallel data for paraphrase evaluation.
\newblock In \emph{Association for Computational Linguistics}, 2011.

\bibitem[Chen et~al.(2024{\natexlab{a}})Chen, Ye, He, Wang, Khashabi, and Yuille]{chenefficient}
Jieneng Chen, Luoxin Ye, Ju He, Zhao-Yang Wang, Daniel Khashabi, and Alan Yuille.
\newblock Efficient large multi-modal models via visual context compression.
\newblock In \emph{Advances in Neural Information Processing Systems}, 2024{\natexlab{a}}.

\bibitem[Chen et~al.(2024{\natexlab{b}})Chen, Li, Dong, Zhang, He, Wang, Zhao, and Lin]{chen2023sharegpt4v}
Lin Chen, Jinsong Li, Xiaoyi Dong, Pan Zhang, Conghui He, Jiaqi Wang, Feng Zhao, and Dahua Lin.
\newblock Sharegpt4v: Improving large multi-modal models with better captions.
\newblock In \emph{Proceedings of the European Conference on Computer Vision}, 2024{\natexlab{b}}.

\bibitem[Chen et~al.(2023)Chen, Hu, Luan, Sun, Changpinyo, Ritter, and Chang]{chen2023can}
Yang Chen, Hexiang Hu, Yi Luan, Haitian Sun, Soravit Changpinyo, Alan Ritter, and Ming-Wei Chang.
\newblock Can pre-trained vision and language models answer visual information-seeking questions?
\newblock In \emph{Proceedings of the Conference on Empirical Methods in Natural Language Processinng}, 2023.

\bibitem[Das et~al.(2017)Das, Kottur, Gupta, Singh, Yadav, Moura, Parikh, and Batra]{das2017visual}
Abhishek Das, Satwik Kottur, Khushi Gupta, Avi Singh, Deshraj Yadav, Jos{\'e}~MF Moura, Devi Parikh, and Dhruv Batra.
\newblock Visual dialog.
\newblock In \emph{Proceedings of the IEEE Conference on Computer Vision and Pattern Recognition}, 2017.

\bibitem[Dubey et~al.(2024)Dubey, Jauhri, Pandey, Kadian, Al-Dahle, Letman, Mathur, Schelten, Yang, Fan, et~al.]{dubey2024llama}
Abhimanyu Dubey, Abhinav Jauhri, Abhinav Pandey, Abhishek Kadian, Ahmad Al-Dahle, Aiesha Letman, Akhil Mathur, Alan Schelten, Amy Yang, Angela Fan, et~al.
\newblock The llama 3 herd of models.
\newblock \emph{arXiv preprint arXiv:2407.21783}, 2024.

\bibitem[Gao et~al.(2021)Gao, Yao, and Chen]{gao2021simcse}
Tianyu Gao, Xingcheng Yao, and Danqi Chen.
\newblock Simcse: Simple contrastive learning of sentence embeddings.
\newblock In \emph{Proceedings of the Conference on Empirical Methods in Natural Language Processinng}, 2021.

\bibitem[Hsieh et~al.(2023)Hsieh, Zhang, Ma, Kembhavi, and Krishna]{hsieh2024sugarcrepe}
Cheng-Yu Hsieh, Jieyu Zhang, Zixian Ma, Aniruddha Kembhavi, and Ranjay Krishna.
\newblock Sugarcrepe: Fixing hackable benchmarks for vision-language compositionality.
\newblock \emph{Advances in Neural Information Processing Systems}, 2023.

\bibitem[Hu et~al.(2023)Hu, Luan, Chen, Khandelwal, Joshi, Lee, Toutanova, and Chang]{hu2023open}
Hexiang Hu, Yi Luan, Yang Chen, Urvashi Khandelwal, Mandar Joshi, Kenton Lee, Kristina Toutanova, and Ming-Wei Chang.
\newblock Open-domain visual entity recognition: Towards recognizing millions of wikipedia entities.
\newblock In \emph{Proceedings of the International Conference on Computer Vision}, 2023.

\bibitem[Huang et~al.(2016)Huang, Ferraro, Mostafazadeh, Misra, Agrawal, Devlin, Girshick, He, Kohli, Batra, et~al.]{huang2016visual}
Ting-Hao Huang, Francis Ferraro, Nasrin Mostafazadeh, Ishan Misra, Aishwarya Agrawal, Jacob Devlin, Ross Girshick, Xiaodong He, Pushmeet Kohli, Dhruv Batra, et~al.
\newblock Visual storytelling.
\newblock In \emph{Proceedings of the Conference of the North American Chapter of the Association for Computational Linguistics}, 2016.

\bibitem[Jia et~al.(2021)Jia, Yang, Xia, Chen, Parekh, Pham, Le, Sung, Li, and Duerig]{jia2021scaling}
Chao Jia, Yinfei Yang, Ye Xia, Yi-Ting Chen, Zarana Parekh, Hieu Pham, Quoc Le, Yun-Hsuan Sung, Zhen Li, and Tom Duerig.
\newblock Scaling up visual and vision-language representation learning with noisy text supervision.
\newblock In \emph{Proceedings of the International Conference on Machine Learning}, 2021.

\bibitem[Jiang et~al.(2023{\natexlab{a}})Jiang, Sablayrolles, Mensch, Bamford, Chaplot, Casas, Bressand, Lengyel, Lample, Saulnier, et~al.]{jiang2023mistral}
Albert~Q Jiang, Alexandre Sablayrolles, Arthur Mensch, Chris Bamford, Devendra~Singh Chaplot, Diego de~las Casas, Florian Bressand, Gianna Lengyel, Guillaume Lample, Lucile Saulnier, et~al.
\newblock Mistral 7b.
\newblock \emph{arXiv preprint arXiv:2310.06825}, 2023{\natexlab{a}}.

\bibitem[Jiang et~al.(2023{\natexlab{b}})Jiang, Huang, Luan, Wang, and Zhuang]{jiang2023scaling}
Ting Jiang, Shaohan Huang, Zhongzhi Luan, Deqing Wang, and Fuzhen Zhuang.
\newblock Scaling sentence embeddings with large language models.
\newblock \emph{arXiv preprint arXiv:2307.16645}, 2023{\natexlab{b}}.

\bibitem[Jiang et~al.(2024{\natexlab{a}})Jiang, Song, Zhang, Huang, Deng, Sun, Zhang, Wang, and Zhuang]{jiang2024e5}
Ting Jiang, Minghui Song, Zihan Zhang, Haizhen Huang, Weiwei Deng, Feng Sun, Qi Zhang, Deqing Wang, and Fuzhen Zhuang.
\newblock E5-v: Universal embeddings with multimodal large language models.
\newblock \emph{arXiv preprint arXiv:2407.12580}, 2024{\natexlab{a}}.

\bibitem[Jiang et~al.(2024{\natexlab{b}})Jiang, Meng, Yang, Yavuz, Zhou, and Chen]{jiang2024vlm2vec}
Ziyan Jiang, Rui Meng, Xinyi Yang, Semih Yavuz, Yingbo Zhou, and Wenhu Chen.
\newblock Vlm2vec: Training vision-language models for massive multimodal embedding tasks.
\newblock \emph{arXiv preprint arXiv:2410.05160}, 2024{\natexlab{b}}.

\bibitem[Kwon et~al.(2023)Kwon, Li, Zhuang, Sheng, Zheng, Yu, Gonzalez, Zhang, and Stoica]{kwon2023efficient}
Woosuk Kwon, Zhuohan Li, Siyuan Zhuang, Ying Sheng, Lianmin Zheng, Cody~Hao Yu, Joseph~E. Gonzalez, Hao Zhang, and Ion Stoica.
\newblock Efficient memory management for large language model serving with pagedattention.
\newblock In \emph{Proceedings of the ACM SIGOPS Symposium on Operating Systems Principles}, 2023.

\bibitem[Lai et~al.(2024)Lai, Tian, Chen, Li, Yuan, Liu, and Jia]{lai2024lisa}
Xin Lai, Zhuotao Tian, Yukang Chen, Yanwei Li, Yuhui Yuan, Shu Liu, and Jiaya Jia.
\newblock Lisa: Reasoning segmentation via large language model.
\newblock In \emph{Proceedings of the IEEE Conference on Computer Vision and Pattern Recognition}, 2024.

\bibitem[Li et~al.(2024{\natexlab{a}})Li, Zhang, Guo, Zhang, Li, Zhang, Zhang, Li, Liu, and Li]{li2024llava-ov}
Bo Li, Yuanhan Zhang, Dong Guo, Renrui Zhang, Feng Li, Hao Zhang, Kaichen Zhang, Yanwei Li, Ziwei Liu, and Chunyuan Li.
\newblock Llava-onevision: Easy visual task transfer.
\newblock \emph{arXiv preprint arXiv:2408.03326}, 2024{\natexlab{a}}.

\bibitem[Li et~al.(2024{\natexlab{b}})Li, Zhang, Zhang, Zhang, Li, Li, Ma, and Li]{li2024llava}
Feng Li, Renrui Zhang, Hao Zhang, Yuanhan Zhang, Bo Li, Wei Li, Zejun Ma, and Chunyuan Li.
\newblock Llava-next-interleave: Tackling multi-image, video, and 3d in large multimodal models.
\newblock \emph{arXiv preprint arXiv:2407.07895}, 2024{\natexlab{b}}.

\bibitem[Li et~al.(2022)Li, Li, Xiong, and Hoi]{li2022blip}
Junnan Li, Dongxu Li, Caiming Xiong, and Steven Hoi.
\newblock Blip: Bootstrapping language-image pre-training for unified vision-language understanding and generation.
\newblock In \emph{Proceedings of the International Conference on Machine Learning}, 2022.

\bibitem[Li et~al.(2023{\natexlab{a}})Li, Li, Savarese, and Hoi]{li2023blip}
Junnan Li, Dongxu Li, Silvio Savarese, and Steven Hoi.
\newblock Blip-2: Bootstrapping language-image pre-training with frozen image encoders and large language models.
\newblock In \emph{Proceedings of the International Conference on Machine Learning}, 2023{\natexlab{a}}.

\bibitem[Li et~al.(2023{\natexlab{b}})Li, Wang, Li, Wang, He, Wang, and Qiao]{li2023unmasked}
Kunchang Li, Yali Wang, Yizhuo Li, Yi Wang, Yinan He, Limin Wang, and Yu Qiao.
\newblock Unmasked teacher: Towards training-efficient video foundation models.
\newblock In \emph{Proceedings of the International Conference on Computer Vision}, 2023{\natexlab{b}}.

\bibitem[Li et~al.(2024{\natexlab{c}})Li, Yuan, Liu, Tang, Wang, Qin, Zhu, and Zhang]{li2024tokenpacker}
Wentong Li, Yuqian Yuan, Jian Liu, Dongqi Tang, Song Wang, Jie Qin, Jianke Zhu, and Lei Zhang.
\newblock Tokenpacker: Efficient visual projector for multimodal llm.
\newblock \emph{arXiv preprint arXiv:2407.02392}, 2024{\natexlab{c}}.

\bibitem[Lin et~al.(2024{\natexlab{a}})Lin, Lee, Shoeybi, Lin, Catanzaro, and Ping]{lin2024mm}
Sheng-Chieh Lin, Chankyu Lee, Mohammad Shoeybi, Jimmy Lin, Bryan Catanzaro, and Wei Ping.
\newblock Mm-embed: Universal multimodal retrieval with multimodal llms.
\newblock \emph{arXiv preprint arXiv:2411.02571}, 2024{\natexlab{a}}.

\bibitem[Lin et~al.(2014)Lin, Maire, Belongie, Hays, Perona, Ramanan, Doll{\'a}r, and Zitnick]{lin2014microsoft}
Tsung-Yi Lin, Michael Maire, Serge Belongie, James Hays, Pietro Perona, Deva Ramanan, Piotr Doll{\'a}r, and C~Lawrence Zitnick.
\newblock Microsoft coco: Common objects in context.
\newblock In \emph{Proceedings of the European Conference on Computer Vision}, 2014.

\bibitem[Lin et~al.(2023)Lin, Chen, Mei, Coca, and Byrne]{lin2023fine}
Weizhe Lin, Jinghong Chen, Jingbiao Mei, Alexandru Coca, and Bill Byrne.
\newblock Fine-grained late-interaction multi-modal retrieval for retrieval augmented visual question answering.
\newblock In \emph{Advances in Neural Information Processing Systems}, 2023.

\bibitem[Lin et~al.(2024{\natexlab{b}})Lin, Mei, Chen, and Byrne]{lin2024preflmr}
Weizhe Lin, Jingbiao Mei, Jinghong Chen, and Bill Byrne.
\newblock {P}re{FLMR}: Scaling up fine-grained late-interaction multi-modal retrievers.
\newblock In \emph{Association for Computational Linguistics}, 2024{\natexlab{b}}.

\bibitem[Liu et~al.(2023{\natexlab{a}})Liu, Li, Wu, and Lee]{liu2024visual}
Haotian Liu, Chunyuan Li, Qingyang Wu, and Yong~Jae Lee.
\newblock Visual instruction tuning.
\newblock In \emph{Advances in Neural Information Processing Systems}, 2023{\natexlab{a}}.

\bibitem[Liu et~al.(2023{\natexlab{b}})Liu, Yao, Zhang, Wang, and Xie]{liu2023zero}
Yikun Liu, Jiangchao Yao, Ya Zhang, Yanfeng Wang, and Weidi Xie.
\newblock Zero-shot composed text-image retrieval.
\newblock In \emph{Proceedings of the British Machine Vision Conference}, 2023{\natexlab{b}}.

\bibitem[Liu et~al.(2021)Liu, Rodriguez-Opazo, Teney, and Gould]{liu2021image}
Zheyuan Liu, Cristian Rodriguez-Opazo, Damien Teney, and Stephen Gould.
\newblock Image retrieval on real-life images with pre-trained vision-and-language models.
\newblock In \emph{Proceedings of the International Conference on Computer Vision}, 2021.

\bibitem[Marino et~al.(2019)Marino, Rastegari, Farhadi, and Mottaghi]{marino2019ok}
Kenneth Marino, Mohammad Rastegari, Ali Farhadi, and Roozbeh Mottaghi.
\newblock Ok-vqa: A visual question answering benchmark requiring external knowledge.
\newblock In \emph{Proceedings of the IEEE Conference on Computer Vision and Pattern Recognition}, 2019.

\bibitem[Mensink et~al.(2023)Mensink, Uijlings, Castrejon, Goel, Cadar, Zhou, Sha, Araujo, and Ferrari]{mensink2023encyclopedic}
Thomas Mensink, Jasper Uijlings, Lluis Castrejon, Arushi Goel, Felipe Cadar, Howard Zhou, Fei Sha, Andr{\'e} Araujo, and Vittorio Ferrari.
\newblock Encyclopedic vqa: Visual questions about detailed properties of fine-grained categories.
\newblock In \emph{Proceedings of the International Conference on Computer Vision}, 2023.

\bibitem[Miech et~al.(2021)Miech, Alayrac, Laptev, Sivic, and Zisserman]{miech2021thinking}
Antoine Miech, Jean-Baptiste Alayrac, Ivan Laptev, Josef Sivic, and Andrew Zisserman.
\newblock Thinking fast and slow: Efficient text-to-visual retrieval with transformers.
\newblock In \emph{Proceedings of the IEEE Conference on Computer Vision and Pattern Recognition}, 2021.

\bibitem[Oord et~al.(2018)Oord, Li, and Vinyals]{oord2018representation}
Aaron van~den Oord, Yazhe Li, and Oriol Vinyals.
\newblock Representation learning with contrastive predictive coding.
\newblock \emph{arXiv preprint arXiv:1807.03748}, 2018.

\bibitem[OpenAI(2023)]{openai2023gpt4v}
OpenAI.
\newblock Gpt-4v(ision) system card, 2023.

\bibitem[Pi et~al.(2023)Pi, Gao, Diao, Pan, Dong, Zhang, Yao, Han, Xu, Kong, et~al.]{pi2023detgpt}
Renjie Pi, Jiahui Gao, Shizhe Diao, Rui Pan, Hanze Dong, Jipeng Zhang, Lewei Yao, Jianhua Han, Hang Xu, Lingpeng Kong, et~al.
\newblock Detgpt: Detect what you need via reasoning.
\newblock \emph{arXiv preprint arXiv:2305.14167}, 2023.

\bibitem[Plummer et~al.(2015)Plummer, Wang, Cervantes, Caicedo, Hockenmaier, and Lazebnik]{plummer2015flickr30k}
Bryan~A Plummer, Liwei Wang, Chris~M Cervantes, Juan~C Caicedo, Julia Hockenmaier, and Svetlana Lazebnik.
\newblock Flickr30k entities: Collecting region-to-phrase correspondences for richer image-to-sentence models.
\newblock In \emph{Proceedings of the International Conference on Computer Vision}, 2015.

\bibitem[Radford et~al.(2021)Radford, Kim, Hallacy, Ramesh, Goh, Agarwal, Sastry, Askell, Mishkin, Clark, et~al.]{radford2021learning}
Alec Radford, Jong~Wook Kim, Chris Hallacy, Aditya Ramesh, Gabriel Goh, Sandhini Agarwal, Girish Sastry, Amanda Askell, Pamela Mishkin, Jack Clark, et~al.
\newblock Learning transferable visual models from natural language supervision.
\newblock In \emph{Proceedings of the International Conference on Machine Learning}, 2021.

\bibitem[Singh et~al.(2025)Singh, Shrivastava, Vatsa, Singh, and Bharati]{singh2024learn}
Jaisidh Singh, Ishaan Shrivastava, Mayank Vatsa, Richa Singh, and Aparna Bharati.
\newblock Learn" no" to say" yes" better: Improving vision-language models via negations.
\newblock In \emph{Winter Conference on Applications of Computer Vision}, 2025.

\bibitem[Sun et~al.(2024)Sun, Wang, Yu, Cui, Zhang, Zhang, and Wang]{sun2024eva}
Quan Sun, Jinsheng Wang, Qiying Yu, Yufeng Cui, Fan Zhang, Xiaosong Zhang, and Xinlong Wang.
\newblock Eva-clip-18b: Scaling clip to 18 billion parameters.
\newblock \emph{arXiv preprint arXiv:2402.04252}, 2024.

\bibitem[Vaze et~al.(2023)Vaze, Carion, and Misra]{vaze2023genecis}
Sagar Vaze, Nicolas Carion, and Ishan Misra.
\newblock Genecis: A benchmark for general conditional image similarity.
\newblock In \emph{Proceedings of the IEEE Conference on Computer Vision and Pattern Recognition}, 2023.

\bibitem[Wang et~al.(2024{\natexlab{a}})Wang, Bai, Tan, Wang, Fan, Bai, Chen, Liu, Wang, Ge, et~al.]{wang2024qwen2}
Peng Wang, Shuai Bai, Sinan Tan, Shijie Wang, Zhihao Fan, Jinze Bai, Keqin Chen, Xuejing Liu, Jialin Wang, Wenbin Ge, et~al.
\newblock Qwen2-vl: Enhancing vision-language model's perception of the world at any resolution.
\newblock \emph{arXiv preprint arXiv:2409.12191}, 2024{\natexlab{a}}.

\bibitem[Wang et~al.(2023)Wang, Chen, Chen, Wu, Zhu, Zeng, Luo, Lu, Zhou, Qiao, et~al.]{wang2024visionllm}
Wenhai Wang, Zhe Chen, Xiaokang Chen, Jiannan Wu, Xizhou Zhu, Gang Zeng, Ping Luo, Tong Lu, Jie Zhou, Yu Qiao, et~al.
\newblock Visionllm: Large language model is also an open-ended decoder for vision-centric tasks.
\newblock In \emph{Advances in Neural Information Processing Systems}, 2023.

\bibitem[Wang et~al.(2022)Wang, Li, Li, He, Huang, Zhao, Zhang, Xu, Liu, Wang, et~al.]{wang2022internvideo}
Yi Wang, Kunchang Li, Yizhuo Li, Yinan He, Bingkun Huang, Zhiyu Zhao, Hongjie Zhang, Jilan Xu, Yi Liu, Zun Wang, et~al.
\newblock Internvideo: General video foundation models via generative and discriminative learning.
\newblock \emph{arXiv preprint arXiv:2212.03191}, 2022.

\bibitem[Wang et~al.(2024{\natexlab{b}})Wang, He, Li, Li, Yu, Ma, Li, Chen, Chen, Wang, et~al.]{wang2023internvid}
Yi Wang, Yinan He, Yizhuo Li, Kunchang Li, Jiashuo Yu, Xin Ma, Xinhao Li, Guo Chen, Xinyuan Chen, Yaohui Wang, et~al.
\newblock Internvid: A large-scale video-text dataset for multimodal understanding and generation.
\newblock In \emph{Proceedings of the International Conference on Learning Representations}, 2024{\natexlab{b}}.

\bibitem[Wang et~al.(2024{\natexlab{c}})Wang, Li, Li, Yu, He, Chen, Pei, Zheng, Xu, Wang, et~al.]{wang2024internvideo2}
Yi Wang, Kunchang Li, Xinhao Li, Jiashuo Yu, Yinan He, Guo Chen, Baoqi Pei, Rongkun Zheng, Jilan Xu, Zun Wang, et~al.
\newblock Internvideo2: Scaling video foundation models for multimodal video understanding.
\newblock In \emph{Proceedings of the European Conference on Computer Vision}, 2024{\natexlab{c}}.

\bibitem[Wei et~al.(2024)Wei, Chen, Chen, Hu, Zhang, Fu, Ritter, and Chen]{wei2023uniir}
Cong Wei, Yang Chen, Haonan Chen, Hexiang Hu, Ge Zhang, Jie Fu, Alan Ritter, and Wenhu Chen.
\newblock Uniir: Training and benchmarking universal multimodal information retrievers.
\newblock In \emph{Proceedings of the European Conference on Computer Vision}, 2024.

\bibitem[Wu et~al.(2021)Wu, Gao, Guo, Al-Halah, Rennie, Grauman, and Feris]{wu2021fashion}
Hui Wu, Yupeng Gao, Xiaoxiao Guo, Ziad Al-Halah, Steven Rennie, Kristen Grauman, and Rogerio Feris.
\newblock Fashion iq: A new dataset towards retrieving images by natural language feedback.
\newblock In \emph{Proceedings of the IEEE Conference on Computer Vision and Pattern Recognition}, 2021.

\bibitem[Xu et~al.(2016)Xu, Mei, Yao, and Rui]{xu2016msr}
Jun Xu, Tao Mei, Ting Yao, and Yong Rui.
\newblock Msr-vtt: A large video description dataset for bridging video and language.
\newblock In \emph{Proceedings of the IEEE Conference on Computer Vision and Pattern Recognition}, 2016.

\bibitem[Xu et~al.(2024)Xu, Liu, Khan, Khan, Zuo, Goh, Feng, et~al.]{xusentence}
Xinxing Xu, Yong Liu, Salman Khan, Fahad Khan, Wangmeng Zuo, Rick Siow~Mong Goh, Chun-Mei Feng, et~al.
\newblock Sentence-level prompts benefit composed image retrieval.
\newblock In \emph{Proceedings of the International Conference on Learning Representations}, 2024.

\bibitem[Yan and Xie(2024)]{yan2024echosight}
Yibin Yan and Weidi Xie.
\newblock {E}cho{S}ight: Advancing visual-language models with {W}iki knowledge.
\newblock In \emph{Findings of the Association for Computational Linguistics: EMNLP}, 2024.

\bibitem[Yuan and Lam(2021)]{yuan2021conversational}
Yifei Yuan and Wai Lam.
\newblock Conversational fashion image retrieval via multiturn natural language feedback.
\newblock In \emph{Proceedings of the International ACM SIGIR Conference on Research and Development in Information Retrieval}, 2021.

\bibitem[Zhai et~al.(2023)Zhai, Mustafa, Kolesnikov, and Beyer]{zhai2023sigmoid}
Xiaohua Zhai, Basil Mustafa, Alexander Kolesnikov, and Lucas Beyer.
\newblock Sigmoid loss for language image pre-training.
\newblock In \emph{Proceedings of the International Conference on Computer Vision}, 2023.

\bibitem[Zhang et~al.(2024{\natexlab{a}})Zhang, Zhang, Dong, Zang, and Wang]{zhang2024long}
Beichen Zhang, Pan Zhang, Xiaoyi Dong, Yuhang Zang, and Jiaqi Wang.
\newblock Long-clip: Unlocking the long-text capability of clip.
\newblock In \emph{European Conference on Computer Vision}, 2024{\natexlab{a}}.

\bibitem[Zhang et~al.(2024{\natexlab{b}})Zhang, Luan, Hu, Lee, Qiao, Chen, Su, and Chang]{zhang2024magiclens}
Kai Zhang, Yi Luan, Hexiang Hu, Kenton Lee, Siyuan Qiao, Wenhu Chen, Yu Su, and Ming-Wei Chang.
\newblock Magiclens: Self-supervised image retrieval with open-ended instructions.
\newblock In \emph{Proceedings of the International Conference on Machine Learning}, 2024{\natexlab{b}}.

\bibitem[Zhao et~al.(2024)Zhao, Ma, Chen, Si, Wu, An, Yu, Zhang, Li, and Chang]{zhao2024ultraedit}
Haozhe Zhao, Xiaojian Ma, Liang Chen, Shuzheng Si, Rujie Wu, Kaikai An, Peiyu Yu, Minjia Zhang, Qing Li, and Baobao Chang.
\newblock Ultraedit: Instruction-based fine-grained image editing at scale.
\newblock In \emph{Advances in Neural Information Processing Systems}, 2024.

\end{thebibliography}
}

% \clearpage
% \setcounter{page}{1}
% \maketitlesupplementary

\onecolumn

{
    \centering
    \Large
    \textbf{LamRA: Large Multimodal Model as Your Advanced Retrieval Assistant}\\
    \vspace{0.5em}Supplementary Material \\
    \vspace{1.0em}
}
\appendix
% {
%   \hypersetup{linkcolor=black}
%   \tableofcontents
% }
% \clearpage

\section{Analysis of Pre-training Dataset Selection}
\label{sec:supp_pretraining_data_selection}

As outlined in Section~\ref{sec:training_for_retrieval} of the main paper, the LamRA-Ret undergoes a two-stage training process to enhance retrieval performance incrementally. These stages consist of the language-only pre-training phase and the multimodal instruction-tuning phase. In this section,
we conduct our pre-training experiments using various types of datasets, which can be categorized into the following groups: (i) datasets containing only image data, (ii) datasets comprising solely text data, (iii) datasets consisting of image-text pairs, and (iv) a combination of pure text data and image-text pair datasets. We utilize MSCOCO and Flickr30K as our evaluation benchmark for the pre-training stage.

\vspace{2pt}
\noindent \textbf{Image-only Dataset.} To construct suitable training data, we utilize the Region-based 100K dataset within UltraEdit~\cite{zhao2024ultraedit}, a region-based image editing dataset. The reference image is employed as the query, the reference image with appropriate data enhancements serves as the positive sample, and the edited image is used as the hard negative sample.

\vspace{2pt}
\noindent \textbf{Language-only Dataset.} We employ the NLI dataset introduced in~\cite{gao2021simcse}, which comprises a series of triplets: (query text, positive text, and hard negative text). The dataset contains 275K samples. In our experiments, we observe that excluding the hard negative text yields improved results. Consequently, in the default experimental setting, we use only the query text and positive text while treating other samples within the same batch as negative samples.

\vspace{2pt}
\noindent \textbf{Image-Text Pairs Dataset.} We use the ShareGPT4V dataset~\cite{chen2023sharegpt4v}, which comprises paired images and long texts, with an initial size of 100K samples. After excluding the COCO images, the final dataset size is reduced to 52K samples. 

\vspace{2pt}
\noindent \textbf{Combined Dataset.} We construct our combined dataset using a combination of the language-only and image-text pairs data.

\vspace{2pt}
\noindent \textbf{Results Analysis.} The experimental results, as presented in Table~\ref{tab:pretraining_experiment}, reveal the following observations: 
(i) Across different types of pertaining data, the retrieval performance consistently improves. 
(ii) Performance fluctuates depending on the data type, with language-only pertaining data yielding the best results. 
(iii) Simply combining different types of pretraining data can unexpectedly degrade performance. 
Based on these findings, we ultimately opt for the language-only pretraining approach.

\begin{table}[htbp]
\small
\label{tab:retrieval_languagebind}
\centering
\resizebox{.8\textwidth}{!}{
\setlength\tabcolsep{1.5mm}
    \begin{tabular}{lcccccccccccc}
        \toprule
         & \multicolumn{6}{c}{\bf{Image Retrieval}} & \multicolumn{6}{c}{\bf{Text Retrieval}} \\
        \cmidrule(r){2-7} \cmidrule(r){8-13}
       \textbf{Dataset Type} & \multicolumn{3}{c}{Flickr30K} & \multicolumn{3}{c}{COCO} & \multicolumn{3}{c}{Flickr30K} & \multicolumn{3}{c}{COCO} \\
        \cmidrule(r){2-4} \cmidrule(r){5-7} \cmidrule(r){8-10} \cmidrule(r){11-13}
         &  R@1 & R@5 & R@10 & R@1 & R@5 & R@10 & R@1 & R@5 & R@10 & R@1 & R@5 & R@10  \\
         \midrule
        No Training & 52.4 & 78.9 & 86.0 & 26.4 & 50.9 & 62.9 & 54.7 & 80.8 & 88.3 & 32.3 & 55.3 & 66.6 \\
        \midrule
        Image-only & 71.3 & 92.0 & 95.2 & 38.9 & 66.2 & 76.6 & 79.3 & 94.0 & 97.3 & 58.0 & 80.0 & 88.0\\
        Language-only & \textbf{80.8} & \textbf{95.3} & \textbf{97.2} & \textbf{53.3} & \textbf{77.0} & \textbf{84.8} & \textbf{91.2} & \textbf{99.0} & \textbf{99.5} & \textbf{67.5} & \textbf{87.1} & \textbf{92.6} \\
        Image-Text Pairs & 72.5 & 90.7 & 94.4 & 46.0 & 71.1 & 79.9 & 88.4 & 97.5 & 99.0 & 65.3 & 85.4 & 91.3 \\
        Combined & 74.1 & 92.5 & 95.8 & 47.2 & 71.8 & 80.4 & 83.9 & 96.0 & 98.4 & 55.1 & 78.9 & 86.3\\
      \bottomrule
      \end{tabular}}
      \caption{\textbf{Performance across various Pre-training Dataset.}}
      \label{tab:pretraining_experiment}
      \vspace{-1em}
\end{table} 

\vspace{2pt}
\noindent \textbf{}

\section{More Implementation Details}

\noindent \textbf{Supplement of Feature Extraction.} As discussed in Section~\ref{sec:feature_extraction} of the main paper, the \texttt{<emb>} token is a newly introduced word added to the vocabulary, primarily serving as a placeholder. 
Essentially, we can use an arbitrary token to replace it, 
since the embedding we use is derived from the hidden state corresponding to the position preceding this token.

\vspace{2pt}
\noindent \textbf{Supplement of Listwise Reranking Method.} 
As discussed in Section~\ref{sec:reranking} of the main paper, the pointwise reranking method assigns a score to each candidate based on the probability of the LMMs outputting \texttt{YES} for that candidate. 
Similarly, the listwise reranking method adopts a similar approach, assigning a reranking score to each candidate according to the probability of the LMM outputting a specific serial number.

\vspace{2pt}
\noindent \textbf{Supplement of Baseline Methods.} 
We utilize the official checkpoints of each baseline model for evaluation. We directly report the results of datasets that have already been evaluated in the corresponding papers. 
Note that, the Coca version of MagicLens~\cite{zhang2024magiclens} has not yet been open-sourced, therefore, we use its CLIP version for our experiments.

\section{Details about M-BEIR Dataset}
\label{sec:supp_m_beir}

We present the details for the M-BEIR benchmark and the corresponding instruction in Table~\ref{tab:mbeir_dataset} and Table~\ref{tab:mbeir_instruction}, respectively. 
It is important to note that the M-BEIR benchmark applies additional processing to the datasets it incorporates, 
which may result in differences from the standard evaluation of individual datasets. 
For instance, the candidate pool of the CIRR dataset in M-BEIR includes training data, which essentially increases the evaluation's difficulty compared to the original CIRR dataset. 
For a more comprehensive understanding of these differences, 
we refer the readers to the original UniIR~\cite{wei2023uniir} paper.

\begin{table}[h]
\centering
\scriptsize % 调整字体大小为更小的脚注级别
\resizebox{.8\textwidth}{!}{  % 控制表格的整体宽度
\setlength{\tabcolsep}{3mm}{
  \begin{tabular}{lllrrrr}
    \toprule
    \textbf{Task} & \textbf{Dataset} & \textbf{Domain} & \textbf{\# Train} & \textbf{\# Dev} & \textbf{\# Test} & \textbf{\# Pool}\\
    \midrule
    \multirow{3}{*}{$q^t \to c^i$} & VisualNews & News & 99K & 20K & 20K & 542K \\
     & MSCOCO & Misc. & 100K & 24.8K & 24.8K & 5K \\
     & Fashion200K & Fashion & 15K & 1.7K & 1.7K & 201K \\
     \midrule
     \multirow{1}{*}{$q^t \to c^t$} & WebQA & Wiki & 16K & 1.7K & 2.4K & 544K \\
     \midrule
     \multirow{2}{*}{$q^t \to (c^i, c^t)$} & EDIS & News & 26K & 3.2K & 3.2K & 1M \\
     & WebQA & Wiki & 17K & 1.7K & 2.5K & 403K \\
     \midrule
     \multirow{3}{*}{$q^i \to c^t$} & VisualNews & News & 100K & 20K & 20K & 537K \\
     & MSCOCO & Misc. & 113K & 5K & 5K & 25K \\
     & Fashion200K & Fashion & 15K & 4.8K & 4.8K & 61K \\
     \midrule
     \multirow{1}{*}{$q^i \to c^i$} & NIGHTS & Misc. & 16K & 2K & 2K & 40K \\
     \midrule
     \multirow{2}{*}{$(q^i, q^t) \to c^t$} & OVEN & Wiki & 150K & 50K & 50K & 676K \\
     & InfoSeek & Wiki & 141K & 11K & 11K & 611K \\
     \midrule
     \multirow{2}{*}{$(q^i, q^t) \to c^i$} & FashionIQ & Fashion & 16K & 2K & 6K & 74K \\
     & CIRR & Misc. & 26K & 2K & 4K & 21K \\
     \midrule
     \multirow{2}{*}{$(q^i, q^t) \to (c^i, c^t)$} & OVEN & Wiki & 157K & 14.7K & 14.7K & 335K \\
     & InfoSeek & Wiki & 143K & 17.6K & 17.6K & 481K \\
     \midrule
    8 tasks & 10 datasets & 4 domains & 1.1M & 182K & 190K & 5.6M \\
    \bottomrule
  \end{tabular}
}
}
\caption{\textbf{Summary of the M-BEIR benchmarks.}
}
\vspace{-2em}
\label{tab:mbeir_dataset}
\end{table}  

\section{Details about Unseen Dataset}
\label{sec:supp_unseen_datasets}

Here, we present the details of the Unseen Dataset in Table~\ref{tab:unseen_dataset}. Many of them are actually adapted from MSCOCO or FashionIQ, however, note that, their captions or query formats are significantly different. 
Therefore, we still treat these datasets as unseen datasets. 
For instance, the captions in Urban1K consist of extended captions generated by GPT-4V~\cite{openai2023gpt4v}, while the query format of CIRCO combines a reference image with a relative caption. 
These differences create a substantial disparity compared to the original COCO dataset.

\begin{table}[h]
\centering
  % 调整字体大小为更小的脚注级别
\resizebox{.8\textwidth}{!}{  % 控制表格的整体宽度
\setlength{\tabcolsep}{1.5mm}{
  \begin{tabular}{lcccc}
    \toprule
    \textbf{Dataset} & \textbf{Image Source} & \textbf{Task} & \textbf{Query Format} & \textbf{Candidate Format}\\
    \midrule
    \multirow{2}{*}{ShareGPT4V} & \multirow{2}{*}{SA-1B} & $q^t \to c^i $ & \texttt{<long text>} & \texttt{<image>} \\
    &  & $q^i \to c^t $ & \texttt{<image>} & \texttt{<long text>} \\
     \midrule
     \multirow{2}{*}{Urban-1K} & \multirow{2}{*}{MSCOCO} & $q^t \to c^i $ & \texttt{<long text>} & \texttt{<image>} \\
    &  & $q^i \to c^t $ & \texttt{<image>} & \texttt{<long text>} \\
     \midrule
     \multirow{2}{*}{Flickr30K} & \multirow{2}{*}{Flickr} & $q^t \to c^i $ & \texttt{<short text>} & \texttt{<image>} \\
    &  & $q^i \to c^t $ & \texttt{<image>} & \texttt{<short text>} \\
    \midrule 
    CIRCO & MSCOCO unlabeled set & $(q^i, q^t) \to c^i $ & \texttt{<image><relative caption>} & \texttt{<image>} \\
    \midrule
    GeneCIS & MSCOCO & $(q^i, q^t) \to c^i $ & \texttt{<image><relative caption>} & \texttt{<image>} \\
    \midrule
    Visual Dialog & MSCOCO & $q^{\text{dialog}} \to c^i $ & \texttt{<Q$_1$><A$_1$>}$\cdots$\texttt{<Q$_\text{j}$><A$_\text{j}$>} & \texttt{<image>} \\
    \midrule 
    Visual Storytelling & Flickr & $ (q^i \oplus q^t) \to c^i $ & \texttt{<text$_1$><image$_1$>}$\cdots$\texttt{<text$_\text{j}$>} & \texttt{<image>} \\
    \midrule
    \multirow{2}{*}{MT-FIQ} & \multirow{2}{*}{FashionIQ} & \multirow{2}{*}{$ (q^i \oplus q^t) \to c^i $} & \multirow{2}{*}{\parbox{7cm}{\texttt{<image$_1$><relative caption$_1$>}$\cdots\\$\texttt{<image$_\text{j}$><relative caption$_\text{j}$>}}} & \multirow{2}{*}{\texttt{<image>}} \\
    & & & & \\
    \midrule
    CC-Neg & CC3M & ITM & \texttt{<image>} & \texttt{<text>} 
    \\
    \midrule
    Sugar-Crepe & MSCOCO & ITM & \texttt{<image>} & \texttt{<text>} \\
    \bottomrule
  \end{tabular}
}
}
\caption{\textbf{Summary of the Unseen Dataset.}
}
\vspace{-2em}
\label{tab:unseen_dataset}
\end{table}
\begin{table}[ht]
\centering
\resizebox{.86\textwidth}{!}{  % 控制表格的整体宽度
\setlength{\tabcolsep}{0.6mm}{
  \begin{tabular}{lll}
    \toprule
    \textbf{Task} & \textbf{Dataset} & \textbf{Instruction} \\
    \midrule
    \multirow{12}{*}{$q^t \to c^i$} & \multirow{4}{*}{VisualNews} & Identify the news-related image in line with the described event. \\
     & & Display an image that best captures the following caption from the news.\\
    & & Based on the caption, provide the most fitting image for the news story. \\
    & & I want you to retrieve an image of this news caption. \\
    \cmidrule(r){2-3}
    & \multirow{4}{*}{MSCOCO} & Find me an everyday image that matches the given caption.\\
    & & Identify the image showcasing the described everyday scene. \\
    & & I want to retrieve an image of this daily life description.\\
    & & Show me an image that best captures the following common scene description.\\
     \cmidrule(r){2-3}
     & \multirow{4}{*}{Fashion200K} & Based on the following fashion description, retrieve the best matching image.\\
     & & Match the provided description to the correct fashion item photo.\\
     & & Identify the fashion image that aligns with the described product.\\
     & & You need to identify the image that corresponds to the fashion product description provided.\\
     \midrule
     \multirow{4}{*}{$q^t \to c^t $} & \multirow{4}{*}{WebQA} & Retrieve passages from Wikipedia that provide answers to the following question.\\
     & & You have to find a Wikipedia paragraph that provides the answer to the question.\\
     & & I want to find an answer to the question. Can you find some snippets that provide evidence from Wikipedia?\\
     & & I'm looking for a Wikipedia snippet that answers this question.\\
     \midrule
     \multirow{8}{*}{$q^t \to (c^i, c^t) $} & \multirow{4}{*}{EDIS} & Find a news image that matches the provided caption.\\
     & & Identify the news photo for the given caption.\\
     & & Can you pair this news caption with the right image?\\
     & & I'm looking for an image that aligns with this news caption. \\
     \cmidrule(r){2-3}
      & \multirow{4}{*}{WebQA} & Find a Wikipedia image that answers this question.\\
     & & Provide with me an image from Wikipedia to answer this question. \\
     & & I want to know the answer to this question. Please find the related Wikipedia image for me. \\
     & & You need to retrieve an evidence image from Wikipedia to address this question. \\
     \midrule
     \multirow{12}{*}{$q^i \to c^t $} & \multirow{4}{*}{VisualNews} & Find a caption for the news in the given photo.\\
     & & Based on the shown image, retrieve an appropriate news caption. \\
     & & Provide a news-related caption for the displayed image. \\
     & & I want to know the caption for this news image. \\
     \cmidrule(r){2-3}
     & \multirow{4}{*}{MSCOCO}                  & Find an image caption describing the following everyday image. \\
     & & Retrieve the caption for the displayed day-to-day image. \\
     & & Can you find a caption talking about this daily life image? \\
     & & I want to locate the caption that best describes this everyday scene image. \\
      \cmidrule(lr){2-3}
     & \multirow{4}{*}{Fashion200K} & Find a product description for the fashion item in the image.  \\
     & & Based on the displayed image, retrieve the corresponding fashion description.  \\
     & & Can you retrieve the description for the fashion item in the image?  \\
     & & I want to find a matching description for the fashion item in this image.  \\
     \midrule
     \multirow{4}{*}{$q^i \to c^i $} & \multirow{4}{*}{NIGHTS} & Find a day-to-day image that looks similar to the provided image.\\
     & & Which everyday image is the most similar to the reference image? \\
     & & Find a daily life image that is identical to the given one. \\
     & & You need to identify the common scene image that aligns most with this reference image. \\
     \midrule 
     \multirow{8}{*}{$(q^i, q^t) \to c^t $} & \multirow{4}{*}{OVEN} & Retrieve a Wikipedia paragraph that provides an answer to the given query about the image.\\
     & & Determine the Wikipedia snippet that identifies the visual entity in the image. \\
     & & I want to find a paragraph from Wikipedia that answers my question about this image. \\
     & & You have to find a Wikipedia segment that identifies this image's subject. \\
     \cmidrule(lr){2-3}
     & \multirow{4}{*}{InfoSeek} & Retrieve a Wikipedia paragraph that provides an answer to the given query about the image.\\
     & & Determine the Wikipedia snippet that matches the question of this image. \\
     & & I want to find a paragraph from Wikipedia that answers my question about this image. \\
     & & You have to find a Wikipedia segment that answers the question about the displayed image. \\
     \midrule 
     \multirow{8}{*}{$(q^i, q^t) \to c^i $} & \multirow{4}{*}{FashionIQ}  & Find a fashion image that aligns with the reference image and style note. \\
     & & With the reference image and modification instructions, find the described fashion look. \\
     & & Given the reference image and design hint, identify the matching fashion image. \\
     & & I'm looking for a similar fashion product image with the described style changes. \\
     \cmidrule(lr){2-3}
     & \multirow{4}{*}{CIRR}  & Retrieve a day-to-day image that aligns with the modification instructions of the provided image. \\
     & & Pull up a common scene image like this one, but with the modifications I asked for. \\
     & & Can you help me find a daily image that meets the modification from the given image? \\
     & & I'm looking for a similar everyday image with the described changes. \\
    \midrule
    \multirow{8}{*}{$(q^i, q^t) \to (c^i, c^t) $} & \multirow{4}{*}{OVEN} & Retrieve a Wikipedia image-description pair that provides evidence for the question of this image. \\
    & & Determine the Wikipedia image-snippet pair that clarifies the entity in this picture. \\
    & & I want to find an image and subject description from Wikipedia that answers my question about this image. \\
    & & I want to know the subject in the photo. Can you provide the relevant Wikipedia section and image? \\
    \cmidrule(lr){2-3}
    & \multirow{4}{*}{InfoSeek} & Retrieve a Wikipedia image-description pair that provides evidence for the question of this image. \\
    & & Determine the Wikipedia image-snippet pair that matches my question about this image. \\
    & & I want to find an image and subject description from Wikipedia that answers my question about this image. \\
    & & I want to address the query about this picture. Please pull up a relevant Wikipedia section and image. \\
    \bottomrule
  \end{tabular}
}
}
\caption{\textbf{Summary of the M-BEIR instructions.}
}
\vspace{-2em}
\label{tab:mbeir_instruction}
\end{table}
\clearpage

\section{Additional Experimental Results}

\subsection{Experimental Results on the M-BEIR in Global Pool Setting}  

The M-BEIR benchmark can be evaluated in two distinct settings: the global pool and the local pool. The key difference between these settings lies in the composition of the candidate pool, which is either constructed from all datasets collectively or restricted to the specific dataset currently under evaluation. 
In the main paper, we report results using the \textbf{local pool setting}. This section provides supplementary evaluation results based on the \textbf{global pool setting}. 

The experimental results are presented in Table~\ref{tab:mbeir_global}. Our method also demonstrates exceptional performance under the global pool setting, 
achieving an average of 12.5 points higher than UniIR-CLIP.

\begin{table*}[h]
% \small
% \setlength{\tabcolsep}{0.1pt}
\centering
\resizebox{\linewidth}{!}{
\begin{tabular}{lc@{\hspace{0.1cm}}c@{\hspace{0.1cm}}c@{\hspace{0.1cm}}c@{\hspace{0.1cm}}c@{\hspace{0.1cm}}c@{\hspace{0.1cm}}c@{\hspace{0.1cm}}c@{\hspace{0.1cm}}c@{\hspace{0.1cm}}c@{\hspace{0.1cm}}c@{\hspace{0.1cm}}c@{\hspace{0.1cm}}c@{\hspace{0.1cm}}c@{\hspace{0.1cm}}c@{\hspace{0.1cm}}c@{\hspace{0.1cm}}c@{\hspace{0.1cm}}}
\toprule
 & \multicolumn{3}{c}{{$q^t \to c^i$}} & {$q^t \to c^t$} & \multicolumn{2}{c}{{$q^t \to (c^i, c^t)$}} & \multicolumn{3}{c}{{$q^i \to c^t$}} & {$q^i \to c^i$} & \multicolumn{2}{c}{{$(q^i, q^t) \to c^t$}} & \multicolumn{2}{c}{{$(q^i, q^t) \to c^i$}} & \multicolumn{2}{c}{{$(q^i, q^t) \to (c^i, c^t)$}} & \\
 \cmidrule(r){2-4} \cmidrule(r){5-5}  \cmidrule(r){6-7} \cmidrule(r){8-10} \cmidrule(r){11-11} \cmidrule(r){12-13} \cmidrule(r){14-15} \cmidrule(r){16-17} 
 Methods & VN  & COCO & F200K & WebQA & EDIS & WebQA & VN & COCO & F200K & NIGHTS & OVEN & InfoS & FIQ & CIRR & OVEN & InfoS & Avg. \\
\cmidrule(r){2-4} \cmidrule(r){5-5}  \cmidrule(r){6-7} \cmidrule(r){8-10} \cmidrule(r){11-11} \cmidrule(r){12-13} \cmidrule(r){14-15} \cmidrule(r){16-17} 
& R@5 & R@5 & R@10 & R@5 & R@5 & R@5 & R@5 & R@5 & R@10 & R@5 & R@5 & R@5 & R@10 & R@5 & R@5 & R@5 & \\
\midrule
\multicolumn{18}{c}{\textit{Supervised - Dual Encoder}} \\
\midrule
$\text{UniIR-BLIP}_{\text{FF}}$~\cite{wei2023uniir} & 23.0 & 75.6 & 25.4 & 79.5 & 50.3 & 79.7 & 21.1 & 88.8 & 27.6 & 33.0 & 38.7 & 19.7 & 28.5 & 51.4 & 57.8 & 27.7 & 45.5  \\
$\text{UniIR-CLIP}_{\text{SF}}$~\cite{wei2023uniir} & 42.6 & 77.9 & 17.8 & 84.7 & 59.4 & 78.8 & 42.8 & 92.3 & 17.9 & 32.0 & 39.2 & 24.0 & 24.3 & 43.9 & 60.2 & 44.6 & 48.9  \\
\midrule
\multicolumn{18}{c}{\textit{Supervised - LMMs}} \\
\midrule
% \addlinespace[0.2em]
% \arrayrulecolor{gray}\hdashline
% \specialrule{0em}{0.5pt}{0.5pt} 
% \largehrulespace 
% LamRA-Ret w/o Pre. & 35.6 & 76.3 & 22.0 & 86.2 & 60.7 & 79.7 & 36.8 & 89.6 & 27.8 & 31.2 & 53.0 & 49.8 & 31.4 & 44.8 & 74.5 & 57.4 & 53.6 \\
% LamRA-Ret w/o Inst. & 16.9 & 73.5 & 10.6 & 80.1 & 45.0 & 60.3 & 14.3 & 81.7 & 12.2 & 28.6 & 24.1 & 31.7 & 7.5 & 28.7 & 36.7 & 27.7 & 36.2\\   
LamRA-Ret & 41.3 & 75.4 & 28.7 & 85.8 & 62.5 & 81.0 & 39.3 & 90.4 & 30.4 & 32.1 & 48.4 & 48.7 & 33.1 & 50.5 & 70.0 & 60.0 & 54.9 \\
LamRA & \textbf{46.9} & \textbf{78.0} & \textbf{32.5} & \textbf{96.5} & \textbf{74.4} & \textbf{87.1} & \textbf{47.6} & \textbf{92.4} & \textbf{36.6} & \textbf{34.2} & \textbf{54.0} & \textbf{58.7} & \textbf{37.4} & \textbf{59.7} & \textbf{72.6} & \textbf{74.0} & \textbf{61.4} \\

\bottomrule
 \end{tabular}
}
\vspace{-5pt}
\caption{\textbf{Comparison with up-to-date state-of-the-arts on M-BEIR test set in global pool setting.} The first row indicates the retrieval task type: $q^t$ for text queries, $q^i$ for image queries, $c^t$ for text candidates, and $c^i$ for image candidates. Abbreviations used include VN for VisualNews, F200K for Fashion200K, InfoS for InfoSeek, and FIQ for FashionIQ. Evaluation standards follow UniIR, with FashionIQ and Fashion200K using Recall@10, while all other evaluations employ Recall@5.} 
\label{tab:mbeir_global}
\vspace{-1em}
\end{table*}

\subsection{Detailed Experimental Results on the Pointwise and Listwise Reranking}

As shown in Table~\ref{tab:supp_point_list}, we present a comprehensive comparison of LamRA-Rank’s pointwise and listwise reranking methods across a range of tasks. 
Both approaches demonstrate enhanced performance in various applications. However, with respect to inference time, the listwise reranking method does not consistently outperform the pointwise approach, particularly when the candidate set includes images. This discrepancy arises because current LMMs often require hundreds of tokens to represent a single image, significantly increasing the context length during listwise reranking. 
The extended context can adversely affect inference speed. 
Ongoing research~\cite{li2024tokenpacker, chenefficient} is actively investigating methods to reduce the number of visual tokens required. We anticipate that, as LMMs continue evolving, the use of LMMs for listwise reranking will become an increasingly prevalent approach.

\begin{table}[ht]
\centering
\scriptsize
\setlength{\tabcolsep}{3mm}
\resizebox{.7\textwidth}{!}{  % 控制表格的整体宽度
  \begin{tabular}{cccccc}
    \toprule
    \multirow{2}{*}{\textbf{Task}} & \textbf{LamRA-Ret} & \multicolumn{2}{c}{\textbf{LamRA-Rank(P)}} & \multicolumn{2}{c}{\textbf{LamRA-Rank(L)}} \\
    \cmidrule(r){2-2} \cmidrule(r){3-4}  \cmidrule(r){5-6}
    & R@1 & R@1 & Time & R@1 & Time \\
    \midrule
    $q^t \to c^i$ & 29.7  & 33.2 & 0.041s & 33.1 & 0.057s \\
      $q^t \to c^t$ & 58.2 & 75.9 & 0.020s & 75.9 & 0.010s\\
    $q^t \to (c^i, c^t)$ & 41.7 & 50.9 & 0.044s & 50.5 & 0.099s\\
    $q^i \to c^t$ & 34.0 & 38.5 & 0.043s & 37.9 & 0.012s\\
      $(q^i, q^t) \to c^i$ & 18.5 & 24.5 & 0.071s & 24.3 & 0.067s\\
    $q^i \to c^i$ & 8.4 & 10.0 & 0.046s & 8.5 & 0.048s\\
      $(q^i, q^t) \to c^t$ & 30.1 & 37.3 & 0.047s & 36.6 & 0.017s\\

      $(q^i, q^t) \to (c^i, c^t)$ & 33.4 & 39.9 & 0.084s & 39.5 & 0.085s \\
    \bottomrule
  \end{tabular}
}
% }
\caption{\textbf{Detailed Comparison of Recall@1 performance and inference costs between pointwise (LamRA-Rank(P)) and listwise (LamRA-Rank(L)) reranking methods on M-BEIR.} Reranking is applied to the top-5 results. Time denotes the per-query inference time cost measured on eight A100 GPUs with a batch size of 32.}
\label{tab:supp_point_list}
\vspace{-1.5em}
\end{table}

\clearpage
\section{Exploration of RAG Applications}
According to the analysis in Section~\ref{sec:sota} of the main paper, the LMM demonstrates significant potential to serve as a universal retriever. 
This raises an important question: 
\textit{Can the retrieval and generative capabilities be integrated within the same LMM?} 

We have conducted experiments on three Knowledge-based Visual Question Answering (KVQA) tasks. 
KVQA requires to first retrieve the relevant documents, then answer the associated questions with the retrieved information. 
Specifically, we train the retrieval and VQA tasks simultaneously using LoRA during the training process. The experimental results, presented in Table~\ref{tab:kvqa}, indicate that the retrieval performance of our method on the three datasets surpasses the current SOTA. Furthermore, the accuracy of answering questions based on the retrieved documents is comparable to or exceeds the current SOTA. These findings demonstrate the feasibility of integrating retrieval and generative capabilities within a single LMM, which we consider a promising direction for future work.

\begin{table}[h]
\centering 
\resizebox{0.7\linewidth}{!}{%
\setlength{\tabcolsep}{10mm}{
\begin{tabular}{@{}lccc@{}}
\toprule
Method                       & OKVQA~\cite{marino2019ok}  & Infoseek~\cite{chen2023can} & E-VQA~\cite{mensink2023encyclopedic}        
\\ \midrule
\multicolumn{4}{c}{\textit{Retrieval (PR@5)}} \\
\midrule
PreFLMR~\cite{lin2024preflmr} & 70.9 & 62.1 & 73.7 \\
Ours & 89.0 & 73.4 & 75.0 \\
\midrule
\multicolumn{4}{c}{\textit{VQA (ACC.)}} \\
\midrule
RA-VQAv2 w/ PreFLMR~\cite{lin2023fine} & 61.9 & 30.7 & 54.5 \\
Ours & 64.3 & 28.8 & 56.2 \\
\bottomrule
\end{tabular}
}}
\caption{\textbf{Comparison with state-of-the-art (SOTA) Methods on KVQA tasks.}}
\label{tab:kvqa}
\end{table}

\section{Limitations \& Future Work}

While our method has demonstrated superior performance over existing models, there remain several limitations that can be further improved. (i) The current implementation of LamRA necessitates training separate sets of LoRA parameters for the retrieval and reranking tasks. In the future, we may explore strategies for jointly training these two tasks simultaneously or consider integrating retrieval training into the SFT stage. (ii) Due to constraints in context length and computation resources, the listwise reranking method of LamRA-Rank currently supports input from 2-5 candidates. Investigating strategies to enable support for a larger number of candidates is an important area for future research. (iii) The current model sizes are restricted to 2B and 7B parameters. Further exploration of LMMs for multimodal information retrieval holds significant potential for improvement.

\section{More Qualitative Results}
\label{sec:supp_qualitative}

\subsection{Successful Cases}

\begin{figure}[h]
    \centering
    \includegraphics[width=\textwidth]{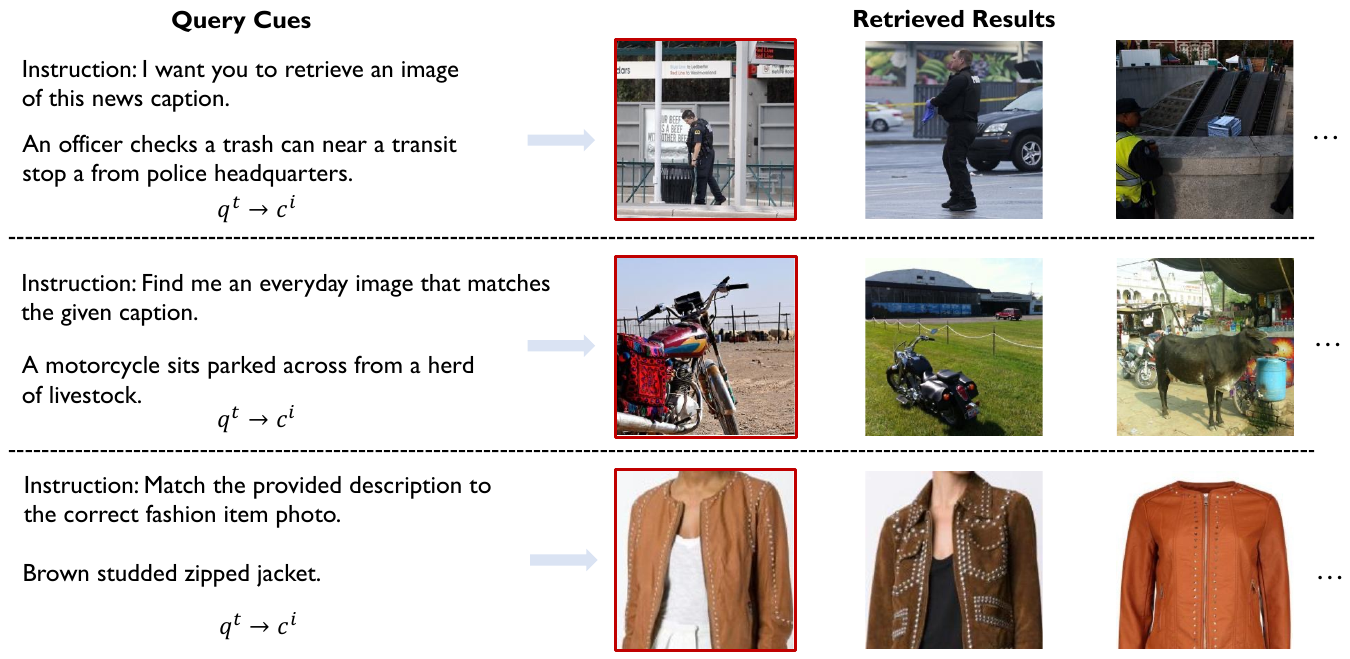}
    \caption{Qualitative examples on text-to-image retrieval task, where the red box marks the ground truth.}
    \label{fig:supp_t2i}
\end{figure}

\begin{figure}[h]
    \centering
    \includegraphics[width=\textwidth]{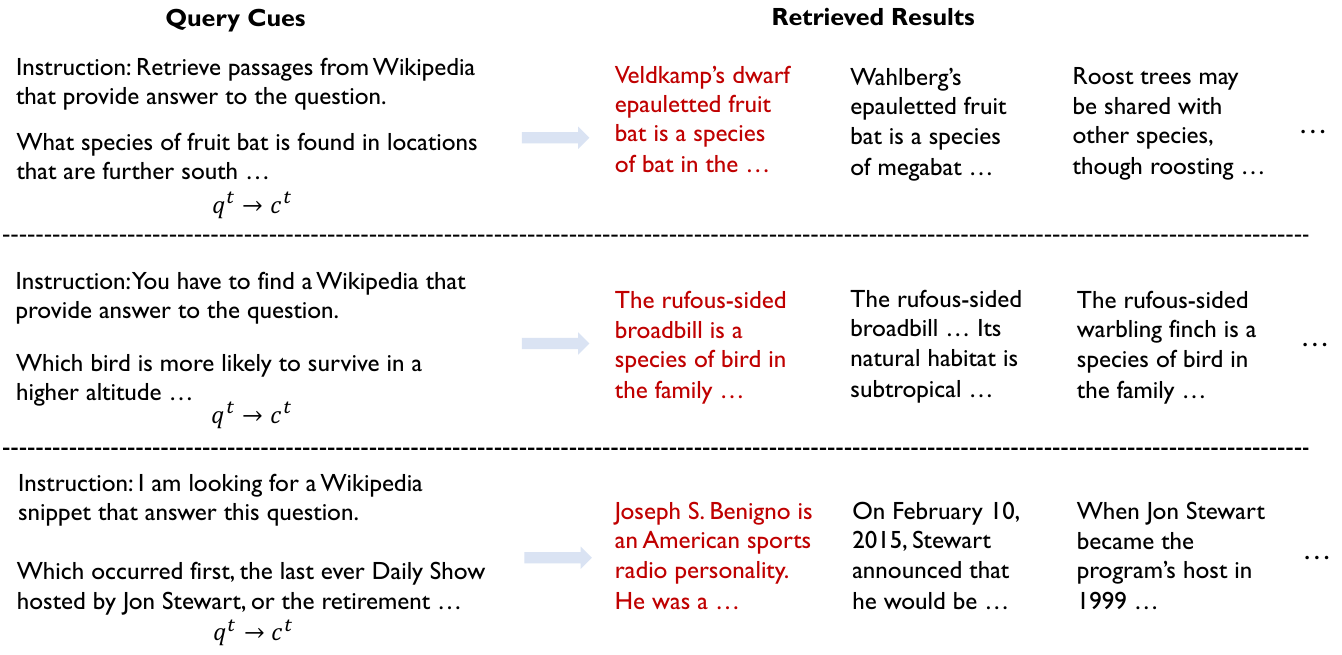}
    \caption{Qualitative examples on text-to-text retrieval task, where the red text marks the ground truth.}
    \label{fig:supp_t2t}
\end{figure}

\begin{figure}[h]
    \centering
    \includegraphics[width=\textwidth]{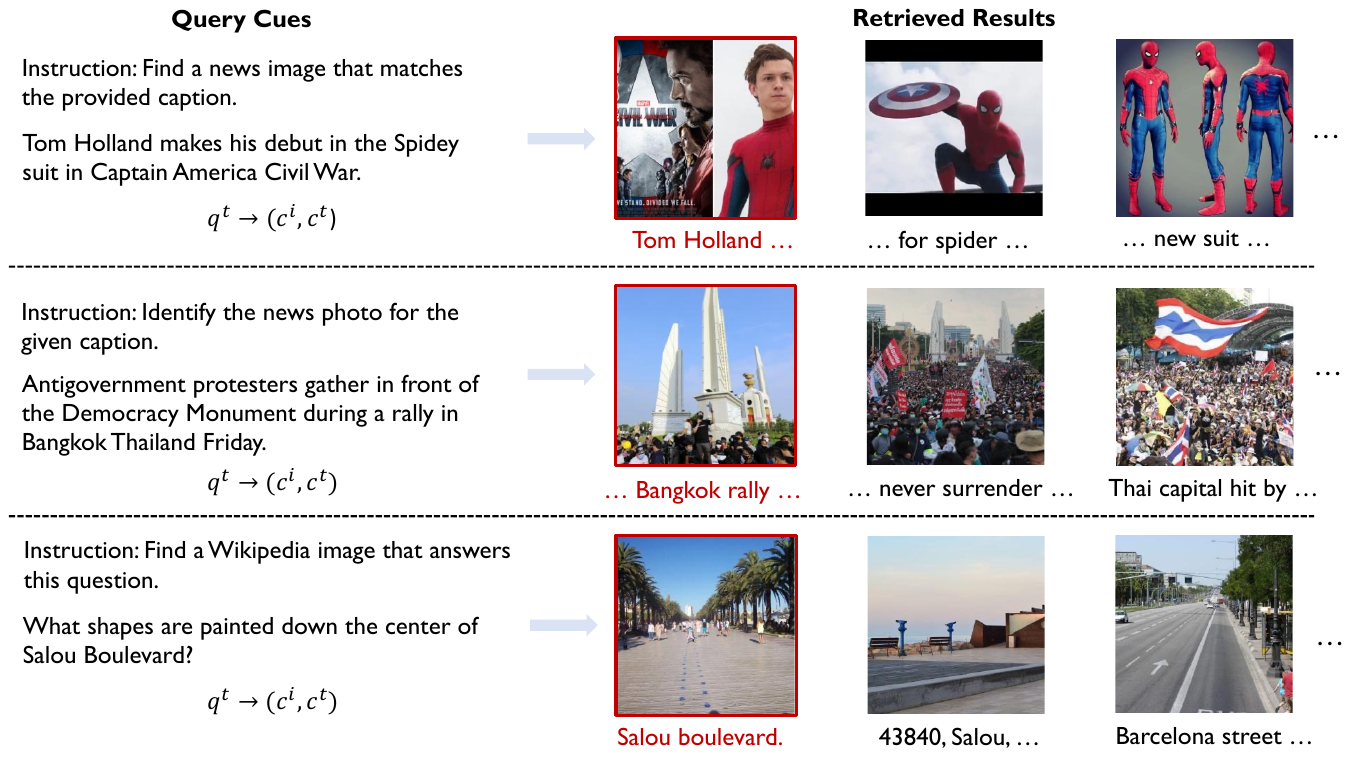}
    \caption{Qualitative examples on text-to-text-image retrieval task, with the ground truth indicated by a red box and red text.}
    \label{fig:supp_t2it}
\end{figure}

\begin{figure}[h]
    \centering
    \includegraphics[width=\textwidth]{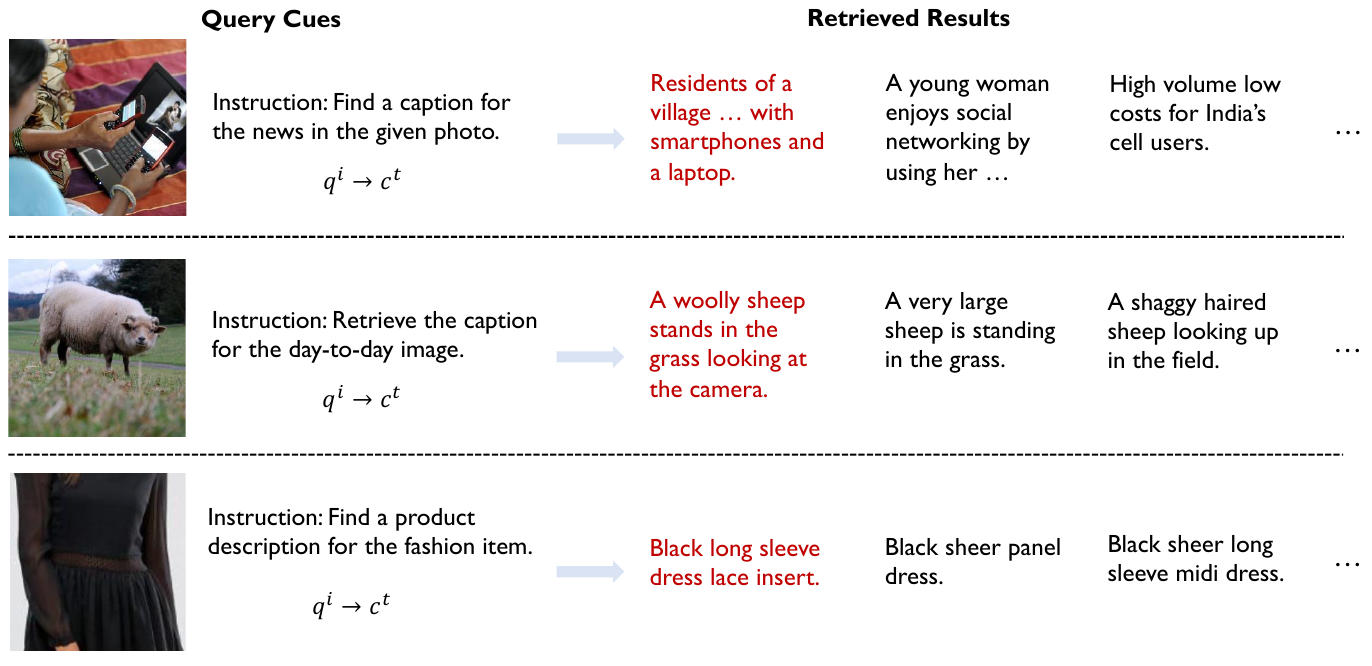}
    \caption{Qualitative examples on image-to-text retrieval task, where the red text marks the ground truth.}
    \label{fig:supp_i2t}
\end{figure}

\begin{figure}[h]
    \centering
    \includegraphics[width=\textwidth]{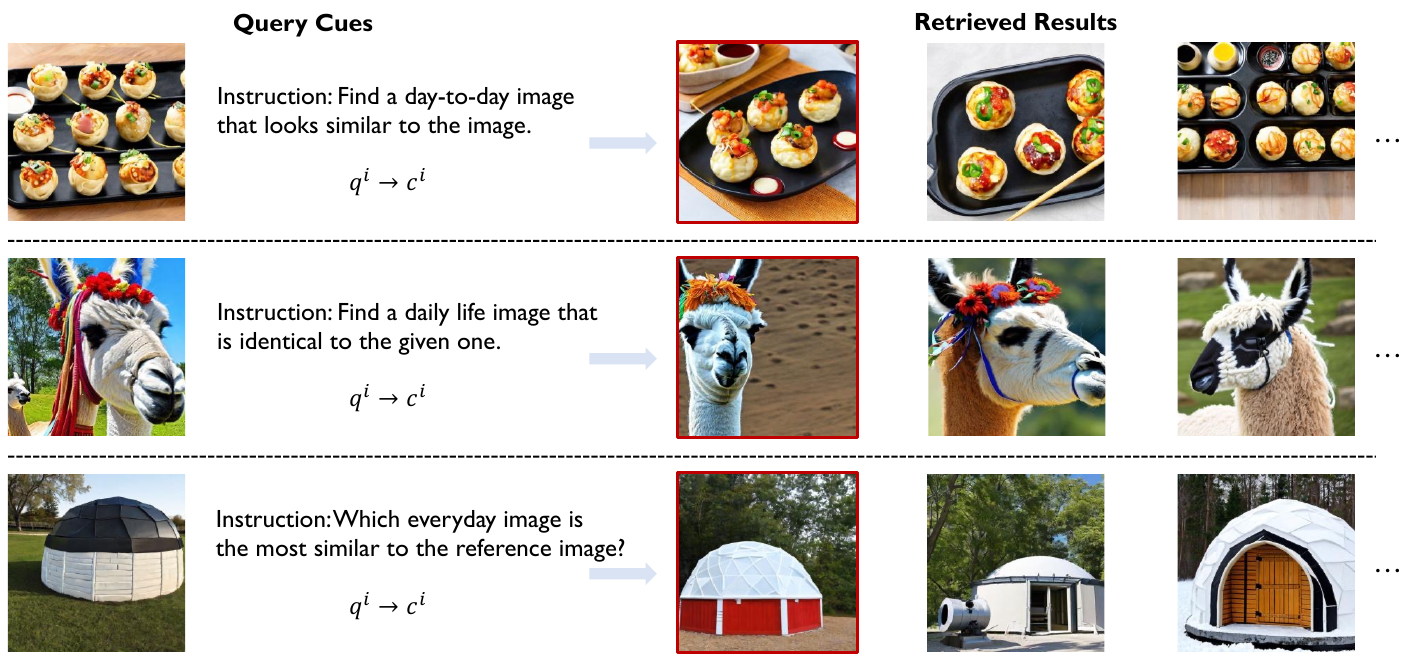}
    \caption{Qualitative examples on image-to-image retrieval task, where the red box marks the ground truth.}
    \label{fig:supp_i2i}
\end{figure}

\begin{figure}[h]
    \centering
    \includegraphics[width=\textwidth]{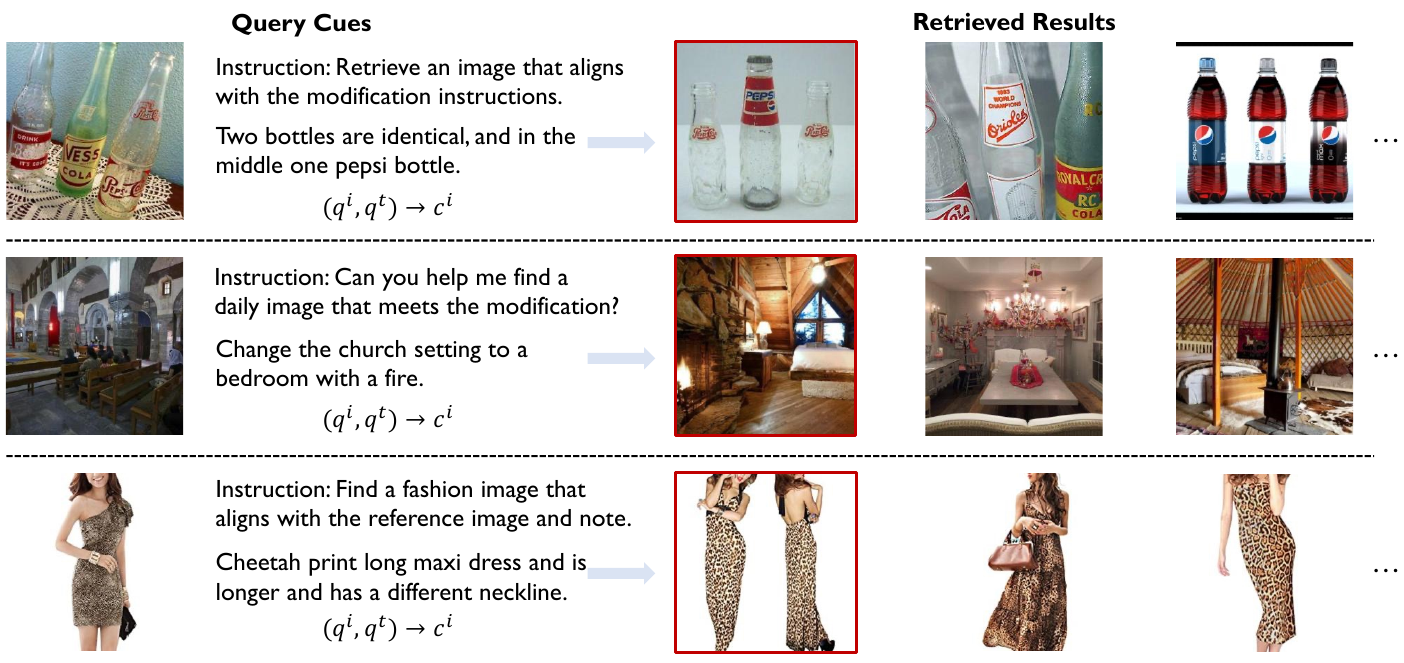}
    \caption{Qualitative examples on text-image-to-image retrieval task, where the red box marks the ground truth.}
    \label{fig:supp_it2i}
\end{figure}

\begin{figure}[h]
    \centering
    \includegraphics[width=\textwidth]{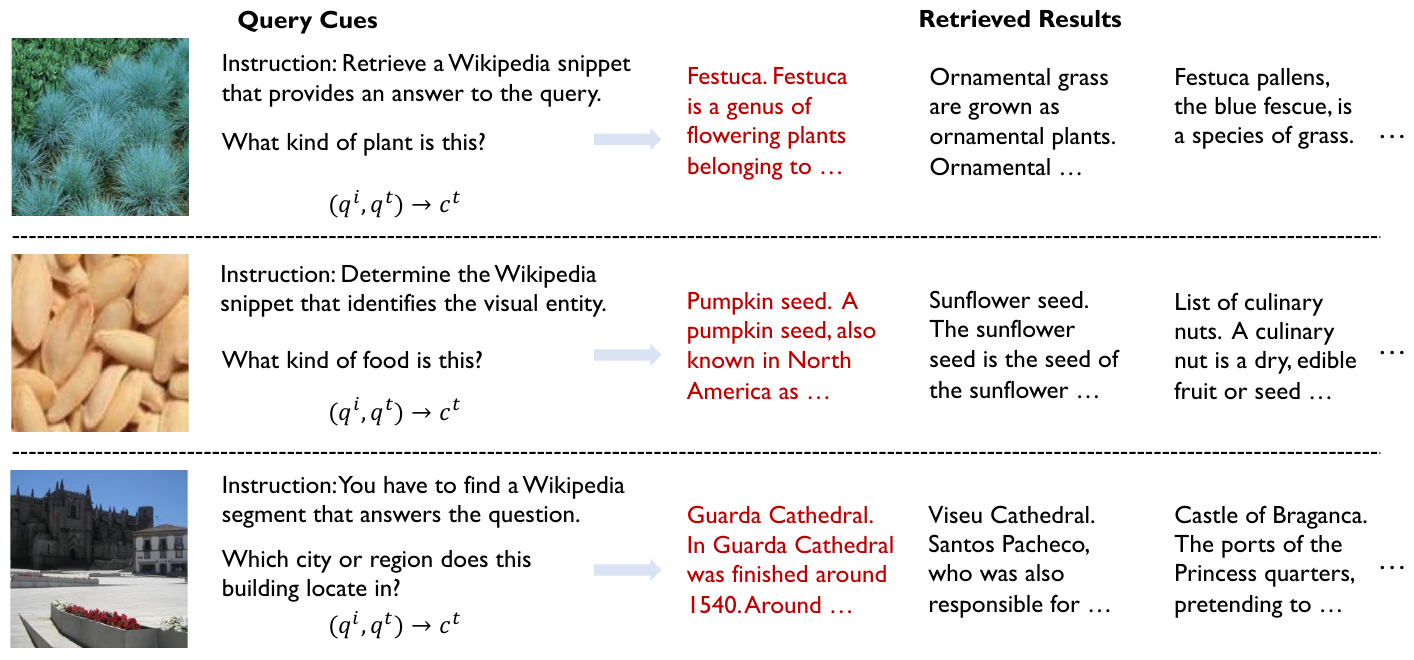}
    \caption{Qualitative examples on text-image-to-text retrieval task, where the red text marks the ground truth.}
    \label{fig:supp_it2t}
\end{figure}

\begin{figure}[h]
    \centering
    \includegraphics[width=\textwidth]{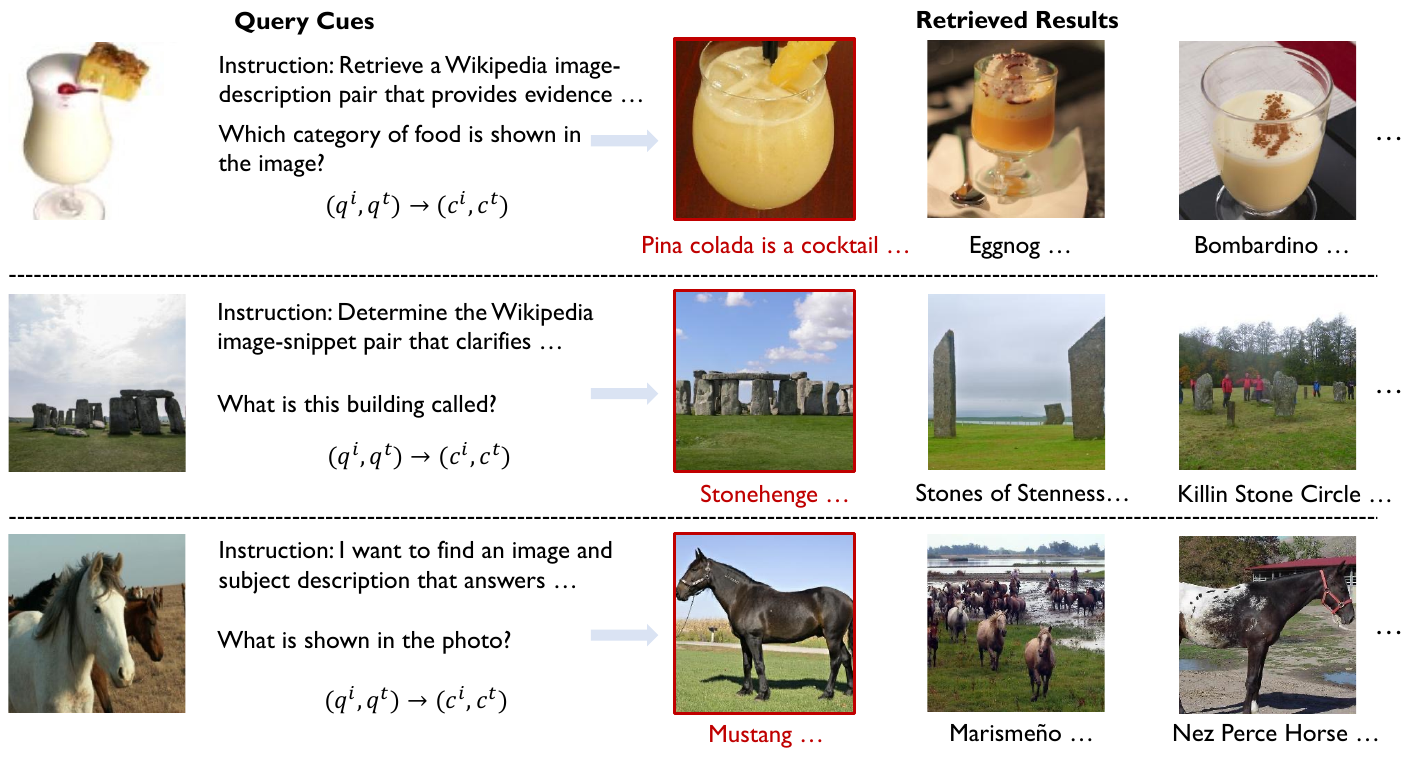}
    \caption{Qualitative examples on text-image-to-text-image retrieval task, with the ground truth indicated by a red box and red text.}
    \label{fig:supp_it2it}
\end{figure}

\begin{figure}[h]
    \centering
    \includegraphics[width=\textwidth]{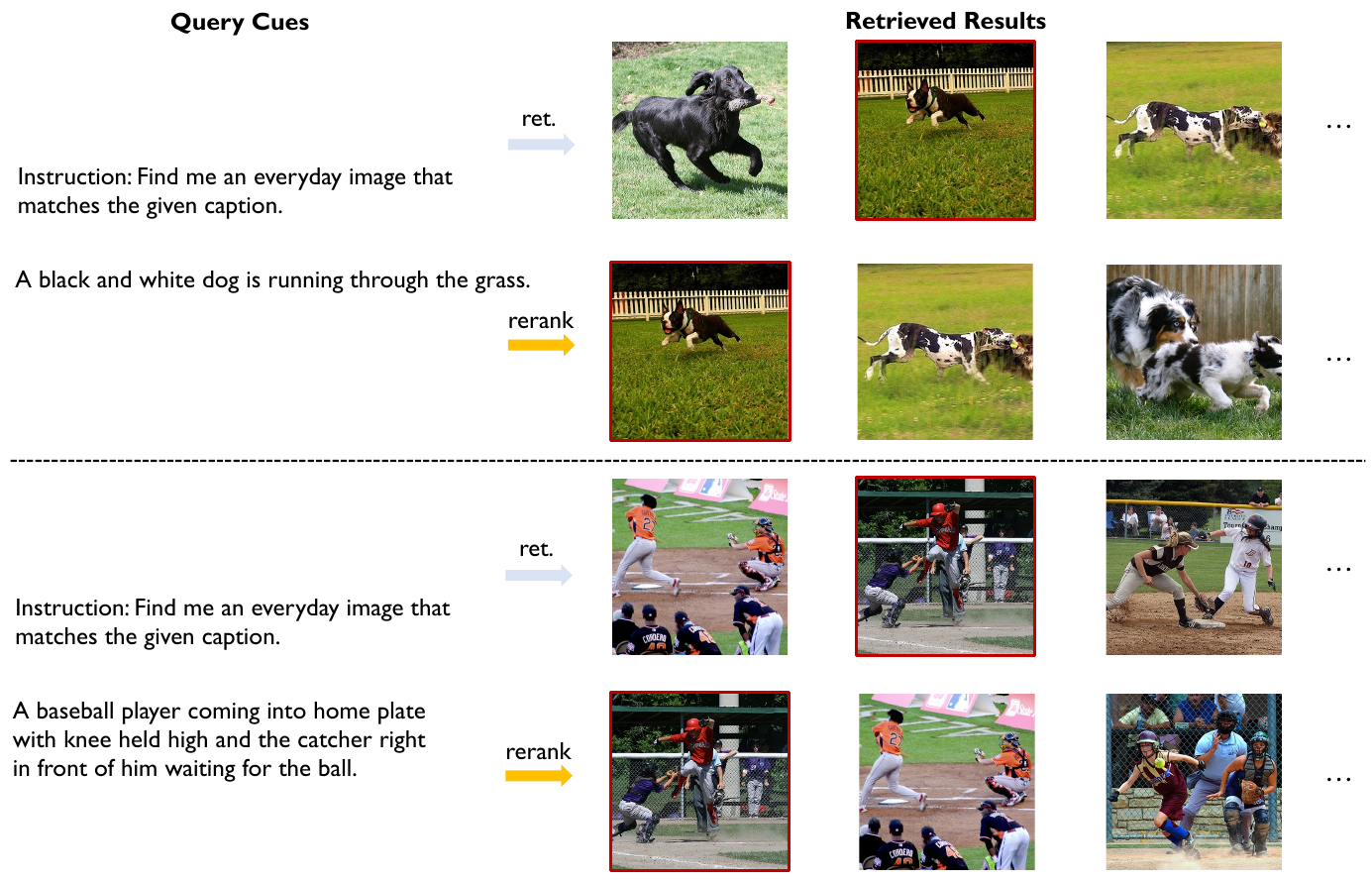}
    \caption{Some examples of retrieval followed by reranking are provided, with the red box indicating the ground truth.}
    \label{fig:supp_rerank}
\end{figure}

In this section, we show additional examples of successful retrieval. As illustrated in Figure~\ref{fig:supp_t2i} through~\ref{fig:supp_rerank}, our method effectively handles a diverse range of retrieval tasks.

\begin{figure}[h]
    \centering
    \includegraphics[width=\textwidth]{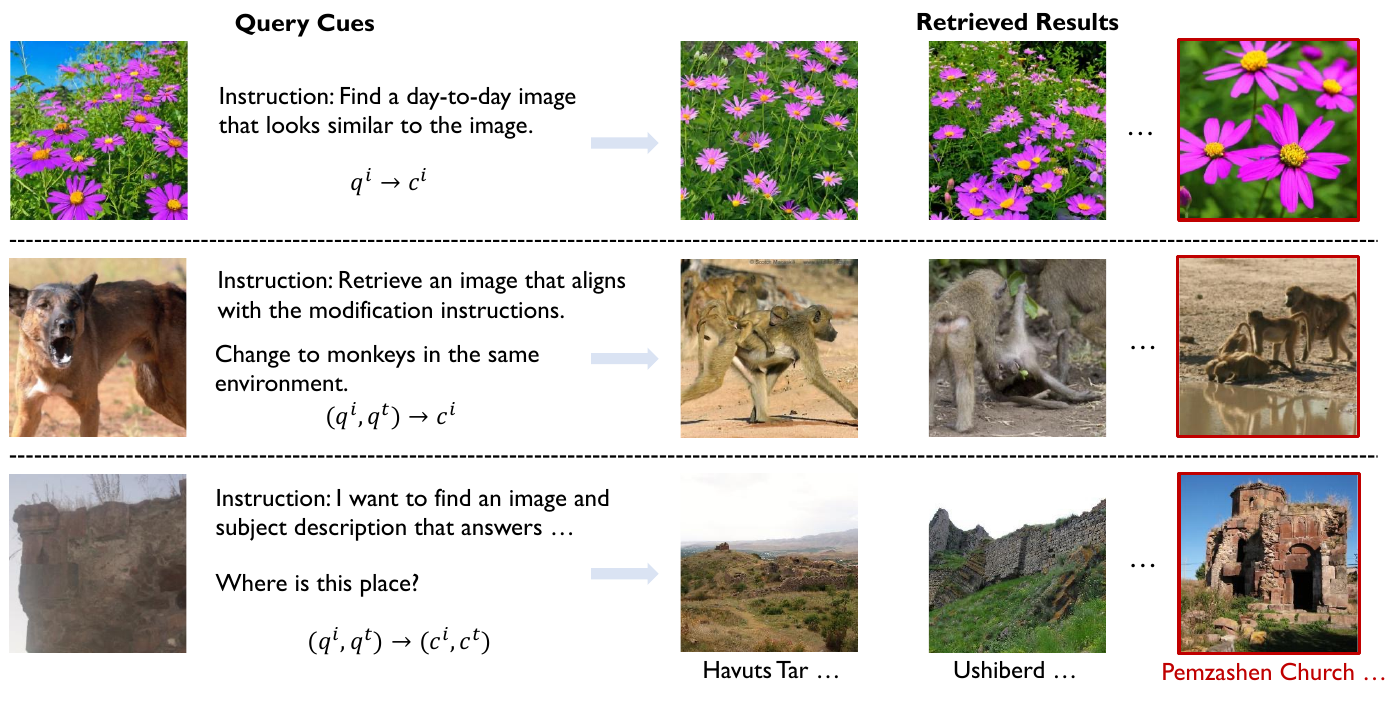}
    \caption{Some failure cases.}
    \label{fig:supp_failure_cases}
\end{figure}

\subsection{Failure Cases}

We present several failure cases in Figure~\ref{fig:supp_failure_cases}. As observed, some failures are attributable to false-negative candidates, while others stem from inherently challenging queries that lead to retrieval failures, as exemplified in the last row.

\clearpage


\end{document}